\documentclass[11pt]{article}
\usepackage{neurodata}

\title{Discovering and Deciphering Relationships Across \\ Disparate Data Modalities}
\author[1,2]{Joshua T. Vogelstein} %,5,6,9
\author[1]{Eric W.~Bridgeford} %7,8
\author[1]{Qing Wang} %7,8
\author[1]{Carey E. Priebe}%,3 \thanks{cep@jhu.edu}}
\author[1]{Mauro Maggioni}%3,4,6\thanks{mauro.maggioni@jhu.edu}}
\author[1,3]{Cencheng Shen} %,2 \thanks{cshen6@jhu.edu}}
\affil[1]{Johns Hopkins University} %Center for Imaging Science,
\affil[2]{Child Mind Institute}
\affil[3]{University of Delaware}

\begin{document}

\maketitle

% @jv: board of reviewing editor options in elife:
% timothy verstynen
% Michael Breakspear
% Alexander Borst
% Peter Latham
% Mark CW van Rossum
% Moritz Helmstaedter
% Stephanie Palmer
% Jörn Diedrichsen
% Jack L Gallant
% David Kleinfeld

\vspace{10pt}

%\begin{abstract}
% eLife version:
\noindent
\textbf{Understanding the relationships between different properties of data, such as whether a genome or connectome has information about disease status, is increasingly important. While existing approaches can test whether two properties are related, they require unfeasibly large sample sizes, and are not interpretable. Our approach, "Multiscale Graph Correlation" (MGC), is a dependence test that juxtaposes disparate data science techniques, including k-nearest neighbors, kernel methods, and multiscale analysis. Other methods typically require double or triple the number samples to achieve the same statistical power as MGC in a benchmark suite including high-dimensional and nonlinear relationships, with dimensionality ranging from 1 to 1000. Moreover, MGC uniquely characterizes the latent geometry underlying the relationship, while maintaining computational efficiency. In real data, including brain imaging and cancer genetics, MGC is the only method that can detects the presence of a dependency and provides guidance for the next experiments to conduct.
% arxiv version:
% Understanding the relationships between different properties of data, such as whether a genome or connectome has information about disease status, is increasingly important. While existing approaches can test whether two properties are related, they require unfeasibly large sample sizes, and lack interpretability. Our approach, ``Multiscale Graph Correlation'' (\Mgc), is a dependence test that juxtaposes previously disparate data science techniques, including k-nearest neighbors, kernel methods (such as support vector machines), and multiscale analysis (such as wavelets).  Other methods typically require double or triple the number samples to achieve the same statistical power as \Mgc~in a  benchmark suite including  high-dimensional and nonlinear relationships---spanning polynomial (linear, quadratic, cubic), trigonometric (sinusoidal, circular, ellipsoidal, spiral),  geometric (square,  diamond, W-shape), and other functions---with dimensionality ranging from 1 to 1000. Moreover, \Mgc~uniquely provides a simple and elegant characterization of the potentially complex latent geometry underlying the relationship, providing  insight while maintaining computational efficiency.  In  real data applications, including brain imaging and cancer genetics,  \Mgc~is the only method that  both detects the presence of a dependency and provides specific guidance for the next experiments to conduct.
}

Identifying the existence of a relationship between a pair of properties or modalities is the critical initial step in data science investigations.
Only if there is a statistically significant relationship does it make sense to try to decipher the nature of the relationship. 
Discovering and deciphering relationships is fundamental, for example,  in high-throughput screening \cite{Zhang1999-js},  precision medicine~\cite{Prescott2013}, machine learning~\cite{HastieTibshiraniBook}, and causal analyses~\cite{Pearl2000}.
One of the first approaches for determining whether two properties are related to---or statistically dependent on---each other is Pearson's Product-Moment Correlation (published in 1895 \cite{Pearson1895}). This seminal paper prompted the development of  entirely new ways of thinking about and quantifying relationships (see \cite{Reimherr2013,JosseHolmes2013} for recent reviews and discussion).
Modern datasets, however, present  challenges for dependence-testing that were not addressed in Pearson's era.
First, we now desire methods that can correctly detect any kind of dependence between all kinds of data, including high-dimensional data (such as 'omics), structured data (such as images or networks), with nonlinear relationships (such as  oscillators), even with very small sample sizes as is common in modern biomedical science.  Second, we desire methods that are interpretable by providing insight into how or why they discovered the presence of a statistically significant relationship. Such insight can be a crucial component of designing the next computational or physical experiment.
% provide insight into the geometry of the underlying relationship---is the relationship linear, quadratic, sinusoidal, exponential, etc.---rather than merely its existence, to help guide further experimentation and analysis.

While many statistical and machine learning approaches have been developed over the last 120 years to combat aspects of the first issue---detecting dependencies---no approach satisfactorily addressed the challenges across all data types, relationships, and dimensionalities.
Hoeffding and Renyi proposed non-parametric tests to address nonlinear but univariate relationships \cite{Hoeffding1948,Renyi1959}.  In the 1970s and 1980s, nearest neighbor style approaches were popularized \cite{Friedman1983,Schilling1986}, but they were sensitive to {algorithm parameters} resulting in poor empirical performance.
``Energy statistics'', and in particular the distance correlation test (\Dcorr), was recently shown to be able to detect any dependency with sufficient observations \cite{SzekelyRizzo2009}, at arbitrary dimensions \cite{SzekelyRizzo2013a}, and structured data \cite{Lyons2013}.
Another set of methods, referred to a ``kernel mean embedding'' approaches, including the Hilbert Schmidt Independence Criterion (\Hsic)~\cite{GrettonGyorfi2010, Muandet2017-od},
have the same theoretical guarantees, which is not surprising given that they are known to be equivalent to energy statistics~\cite{SejdinovicEtAl2013,Shen2018-st}.
Both energy statistics methods and kernel methods perform very well empirically  with a relatively small sample size  on high-dimensional linear data, whereas another test (Heller, Heller, and Gorfine's test, \Hhg)~\cite{HellerGorfine2013} perform well on low-dimensional nonlinear data.
But no test performs particularly well on high-dimensional nonlinear data with typical sample sizes, which characterizes a large fraction of real data challenges in the current big data era.

Moreover, to our knowledge, no method for detecting dependence has even addressed the issue of interpretability. 
On the other hand, much effort has been devoted to characterizing ``point cloud data'', that is, summarizing certain global properties in unsupervised settings (for example, having genomics data, but no disease data).  
Classic examples of such approaches include Fourier~\cite{Bracewell1986-bq} and wavelet analysis~\cite{Daubechies1992-lc}.
More recently,    topological and geometric data analysis compute properties of graphs, or even higher order simplices~\cite{Edelsbrunner2009-zx}.  Such methods build multiscale characterization of the samples, much like recent developments in harmonic analysis~\cite{Coifman2006-pu,Allard2012}.  However, these tools typically lack statistical guarantees under noisy observations, and are often quite computationally burdensome.

We surmised that both (i) empirical performance in high-dimensional, nonlinear, low-sample size settings, and (ii) providing insight into the discovery process,  could be satisfactorily addressed via extending existing dependence tests to be
 \emph{adaptive} to the data \cite{zhang2012adaptive}.  Existing tests rely on a fixed \emph{a priori} selection of an algorithmic parameter, such as the kernel bandwidth \cite{gretton2006kernel}, intrinsic dimension \cite{Allard2012}, and/or local scale \cite{Friedman1983,Schilling1986}. Indeed, the Achilles Heel of manifold learning has been the requirement to manually choose these parameters \cite{levina2004maximum}. 
Post-hoc cross-validation is often used to make these methods effective adaptive, but doing so adds an undesirable computational burden, and may weaken or destroy any statistical guarantees.
There is therefore a need for statistically valid and computationally efficient adaptive methods.

%\begin{wrapfigure}{r}{0pt}
\begin{figure}
\centering
	\includegraphics[width=1.0\textwidth,trim={0cm 0cm 0cm 0cm},clip]
{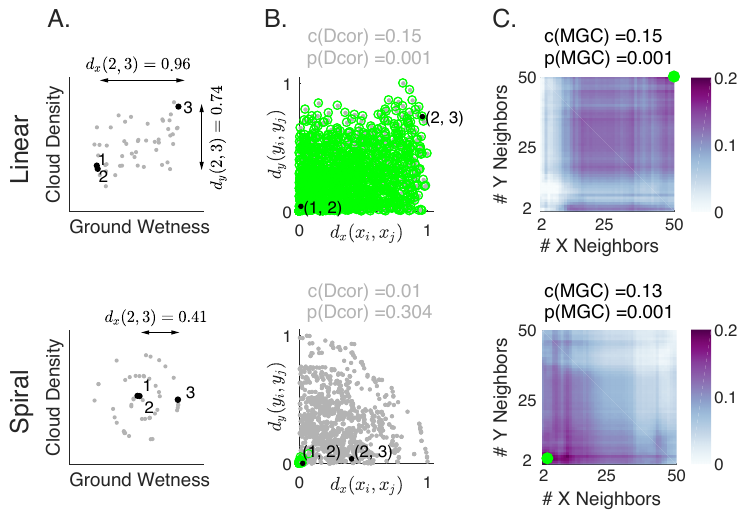}
\caption{
Illustration of Multiscale Graph Correlation (\Mgc)  simulating cloud density ($x_i$) and grass wetness ($y_i$).
We present two different relationships:  linear (top) and nonlinear spiral  (bottom; see Appendix \ref{appen:function} for simulation details).
(\textbf{A})  Scatterplots of the raw data using  $50$ pairs of samples for each scenario.   Samples $1$, $2$, and $3$ (black) are highlighted; arrows show $x$ distances between these pairs of points while their $y$ distances are almost $0$.
(\textbf{B}) Scatterplots of all pairs of distances comparing $x$ and $y$ distances. Distances are linearly correlated in the linear relationship, whereas they are not in the spiral relationship.
\Dcorr~uses all distances (gray dots) to compute its test statistic and p-value, whereas \Mgc~chooses the local scale and then uses only the local distances (green dots). % whose scale is chosen in the next step.
(\textbf{C})  Heatmaps characterizing the strength of the generalized correlation at all possible scales (ranging from $2$ to $n$ for both $x$ and $y$).
For the linear relationship, the global scale is optimal, and is the scale that \Mgc~selects, resulting in a p-value identical to \Dcorr.
For the nonlinear relationship, the optimal scale is local in both $x$ and $y$, so \Mgc~achieves a far larger test statistic, and a correspondingly smaller and significant p-value.
Thus, \Mgc~uniquely detects dependence and characterizes the geometry in both relationships.
} \label{f:newschem}
\end{figure}
%\end{wrapfigure}

To illustrate the importance of adapting to different kinds of relationships, consider a  simple illustrative example:  investigate the relationship between cloud density and grass wetness. If this relationship were approximately linear, the data might look like those in Figure \ref{f:newschem}{\color{magenta}A} (top).
On the other hand, if the relationship were nonlinear---such as a  spiral---it might look like those in Figure \ref{f:newschem}{\color{magenta}A} (bottom).
Although the relationship between clouds and grass is unlikely to be spiral, spiral relationships are prevalent in nature and mathematics (for example, shells, hurricanes, and galaxies), and are canonical in evaluations of manifold learning techniques \cite{Lee07a}, thereby motivating its use here.

% \don{what  is a neighbor, what is geometry, whole introduction feels missing}
Under the linear relationship (top panels), when a pair of observations are close to each other in cloud density, they also tend to be close to each other in grass wetness (for example, observations 1 and 2 highlighted in black in Figure \ref{f:newschem}{\color{magenta}A}, and distances between them in Figure \ref{f:newschem}{\color{magenta}B}).
Similarly, when a pair of observations are far from each other in cloud density, they also tend to be far from each other in grass wetness (see for example, distances between observations 2 and 3).
On the other hand, consider the nonlinear (spiral) relationship (bottom panels).  Here, when a pair of observations are close to each other in cloud density, they also tend to be close to each other in grass wetness (see points 1 and 2 again).
However,  the same is not true for large distances  (see points 2 and 3).
Thus, in the linear relationship, the distance between every pair  of points is informative with respect to the relationship, while under the nonlinear relationship, only a subset of the distances are.

For this reason, we juxtapose nearest neighbor methods with  distance/kernel methods.  Specifically, 
for each point, we find its $k$-nearest neighbors for one property (e.g., cloud density), and its $l$-nearest neighbors for the other property (e.g., grass wetness);  we call the pair $(k,l)$ the ``scale''.  
\emph{A priori}, however, we do not know which scales will be most informative.  Therefore, leveraging recent ideas from multiscale analysis, we efficiently compute the distances for all scales.
The test statistics, described in detail below, summarize the correlations between distances at each  scale  (Figure \ref{f:newschem}{\color{magenta}C}), illustrating which scales are relatively informative about the relationship.  
The key, therefore, to successfully discover and decipher relationships between disparate data modalities is to adaptively determine  which scales are the most informative. 
Doing so not only provides an estimate of whether the modalities are related, but also provides insight into how the determination was made. 
This is especially important in high-dimensional data, where simple visualizations do not reveal relationships to the unaided human eye.
%
%:  the ``jointly local'' distances, that is, those that are close along both dimensions or properties. 
%The key, therefore, to successfully determining the presence and geometry of a relationship is to adaptively estimate how ``local'' to look, that is, to determine the optimal ``scale''.

%By characterizing the strength of dependence at all scales, one can obtain both (1) an understanding of the geometry underlying the relationship, and (2) determine which distances are sufficiently close to warrant inclusion for assessing overall dependence, thereby improving sensitivity and specificity of the test.

Our method, ``Multiscale Graph Correlation'' (\Mgc, pronounced ``magic''), generalized and extends  previously proposed pairwise comparison-based approaches by adaptively estimating the informative scales for any relationship~---~linear or nonlinear, low-dimensional or high-dimensional, unstructured or structured---in a computationally efficient and statistically consistent fashion. This adapative nature of \Mgc~effectively guarantees equally good or better statistical performance compared to existing global methods.
Moreover, the dependence strength across all  scales is informative about how \Mgc~determined the existence of a statistical relationship,  therefore providing further guidance for subsequent experimental or analytical steps. 
\Mgc~is thus a hypothesis-testing and insight-providing approach that builds on recent developments in manifold and kernel learning (operating on pairwise comparisons) by combining them with complementary developments in nearest-neighbor search, and multiscale analyses. It is this union of  disparate disciplines spanning data science that enables improved theoretical and empirical performance. We provide an R package called \Mgc~distributed on the Comprehensive R Archive Network (CRAN) to enable others to use this method for a wide variety of applications~\cite{Bridgeford2018-la}.
% Our open source implementation\footnotemark\footnotetext{In both MATLAB and R from our website, \website.}

\subsection*{The Multiscale Graph Correlation Procedure}

\Mgc~is a multi-step procedure to discover and decipher dependencies across disparate data modalities or properties, as follows  (see Appendix~\ref{appen:mgc} and~\cite{mgc2} for details):
%``\Mgc-Map'' is the matrix of all local generalized correlations).
\begin{enumerate}
	\item  Compute two  distance matrices, one consisting of distances between all pairs of one property (e.g.,  cloud densities, entire genomes or connectomes) and the other consisting of distances between all pairs of the other property  (e.g., grass wetnesses or disease status). Then center each matrix (by subtracting its overall mean, the column-wise mean from each column, and the row-wise mean from each row).		Call the resulting n-by-n matrices ${A}$ and ${B}$.
	\item Compute  the $k$-nearest neighbor graphs for one property, and the $l$-nearest neighbor graph for the other property, for all possible values of $k$ and $l$.  Let $\{{G}_k\}$ and $\{{H}_l\}$ be the nearest neighbor graphs for all $k,l = 1,\ldots, n$, where ${G}_k(i,j)=1$ indicates that $A(i,j)$ is within the $k$ smallest values of the $i^{th}$ row of $A$.  Note that this yields  $n^2$ binary n-by-n matrices, 
	\item Estimate the  local generalized correlation, that is, the correlation between distances restricted to only the $(k,l)$ neighbors by summing the products of these matrices, $t_{kl} = \sum_{ij} {A}(i,j) {G}_k(i,j)  {B}(i,j)  {H}_l(i,j)$, for all values of $k$ and $l$. % , that is, the  correlation of the distances of the $(k,l)$ nearest neighbors.
	\item  Estimate the optimal local generalized correlation, $\hat{t}_{*}$ by finding the smoothed maximum of all local generalized correlations, $t_{kl}$. Smoothing avoids biases and provides \Mgc~with better finite-sample performance and stronger theoretical guarantees.
	\item   Determine whether the relationship is significantly dependent---that is, whether $\hat{t}_{*}$ is more extreme than expected under the null---via a permutation test. The permutation procedure repeats steps 1-4 on each permutation, thereby eliminating the multiple hypothesis testing problem by only computing one overall p-value, rather than one p-value per scale, ensuring that it is a valid test (meaning that the false positive rate is properly controlled at the specified type I error rate).
\end{enumerate}

%We note several clarification points here.
%The first step of \Mgc~is the same as many non-parametric methods. However, global methods then compute the ``generalized correlation'', which is simply the correlation between all distances (see Appendix \ref{appen:mgc} for details on the global methods).
%In contrast, \Mgc~computes \Mgc-Map, which characterizes the geometry of the relationship (Figure~\ref{f:newschem}{\color{magenta}C}).
%Note how different the  \Mgc-Maps look for linear versus spiral relationships.
%Similarly, the optimal scales are quite different for the linear versus spiral relationship: for the linear relationship the optimal scale is global, but for the spiral it is quite local. The green dots in Figure \ref{f:newschem}{\color{magenta}B} show the set of distances amongst the $(k,l)$ nearest neighbors that \Mgc~selected for these particular simulations. And the green dot in Figure \ref{f:newschem}{\color{magenta}C} shows \Mgc's estimated optimal scale.
%
%Finally, 
Running \Mgc~is straightforward---simply input $n$ paired samples of two measured properties, or two dissimilarity matrices of size $n \times n$.
%Our open source implementation\footnotemark\footnotetext{In both MATLAB and R from our website, \website.} 
Computing all local generalized correlations, the test statistic, and p-value  requires $O(n^2 \log n)$ time, which is about the same running -time complexity as other methods. 
%Because our open source implementation is parallel, in practice, depending on the number of cores, it can run faster than competing methods.

%\afterpage{

\subsection*{\Mgc~Typically Requires Substantially Fewer Samples to Achieve the Same Power Across  All Dependencies and Dimensions}

When, and to what extent, does \Mgc~outperform other approaches, and when does it not?
To address this question, we formally pose the following hypothesis test (see Appendix \ref{appen:mgc} for details):
\begin{align*}
H_0\!:& \; X \text{ and } Y \text{ are independent} \\
H_A\!:& \; X \text{ and } Y \text{ are \emph{not} independent}.
\end{align*}
The standard criterion for evaluating statistical tests is to compute the probability that it correctly rejects a false null hypothesis, the testing power, at a given type 1 error level.
In a complementary  manuscript~\cite{mgc2}, we established the theoretical properties of \Mgc, including proving its validity and universal consistency for dependence testing against all distributions of finite second moments, 
meaning that it will reject false null hypotheses for  any dependency with enough samples.
. %, and demonstrated its finite-sample advantage over distance correlation.
%In a complementary theoretical manuscript currently under review, we define ``Oracle \Mgc'', which is the same as \Mgc~proposed here, but it knows (rather than estimates) the optimal scale~\cite{mgc2}.   In that work, we prove that for any global generalized correlation for which one can devise a multiscale extension (such as \Mantel~or \Dcorr), Oracle \Mgc~statistically dominates its corresponding global variant, meaning that for all sample sizes, dimensions, and relationships, Oracle \Mgc~achieves the same or higher power.

Here, we address the empirical performance of \Mgc~as compared with multiple popular tests:
(i) \Dcorr, a popular approach from the statistics community~\cite{SzekelyRizzoBakirov2007,SzekelyRizzo2009},
(ii) \Mcorr, a modified version of \Dcorr~designed to be unbiased for sample data \cite{SzekelyRizzo2013a},
(iii) \Hhg, a distance-based test that is very powerful for detecting low-dimensional nonlinear relationships \cite{HellerGorfine2013}.
(iv) \Hsic, a kernel test \cite{GrettonGyorfi2010} which is equivelant to \Dcorr~but using a different kernel~\cite{Shen2018-st},
(v) \Mantel, which is historically widely used in biology and ecology \cite{Mantel1967}.
(vi) RV coefficient \cite{Pearson1895, JosseHolmes2013}, which is a multivariate generalization of \Pearson's product moment correlation whose test statistic is the sum of the trace-norm of the cross-covariance matrix, and
(vii)  the \CCA~method, which is the largest (in magnitude) singular value of the cross-covariance matrix, and can be viewed as a different generalization of \Pearson~in high-dimensions that is more appropriate for sparse settings~\cite{Hotelling1936, WittenTibshiraniHastie2009,WittenTibshirani2011,TenenhausTenenhaus2011}.
Note that while we focus on high-dimensional settings, Appendix~\ref{appen:figs} shows further results in one-dimensional settings, also comparing to a number of tests that are limited to one dimension, including:
(viii) \Pearson's product moment correlation,
(ix) \Spearman's rank correlation \cite{Spearman1904},
(x) \Kendall's tau correlation \cite{KendallBook}, and
(xi) \Mic~\cite{Reshef2011}.
Under the regularity condition that the data distribution has finite second moment, the first four tests are universally consistent, whereas the other tests are consistent only for linear or monotone relationships.
% and \Mic~lacks any theoretical guarantee.
%Moreover, \Pearson, \RV~and \CCA~are grouped together since they are intrinsically the same in idea and performance; \Spearman~and \Kendall~are both rank correlations and also perform the same, both of which are not applicable to multivariate data just like \Mic.

\begin{figure}[htbp]
\includegraphics[width=1.0\linewidth]
% [width=1.0\textwidth,trim={0 0.5cm 3.5cm 0.5cm},clip]
{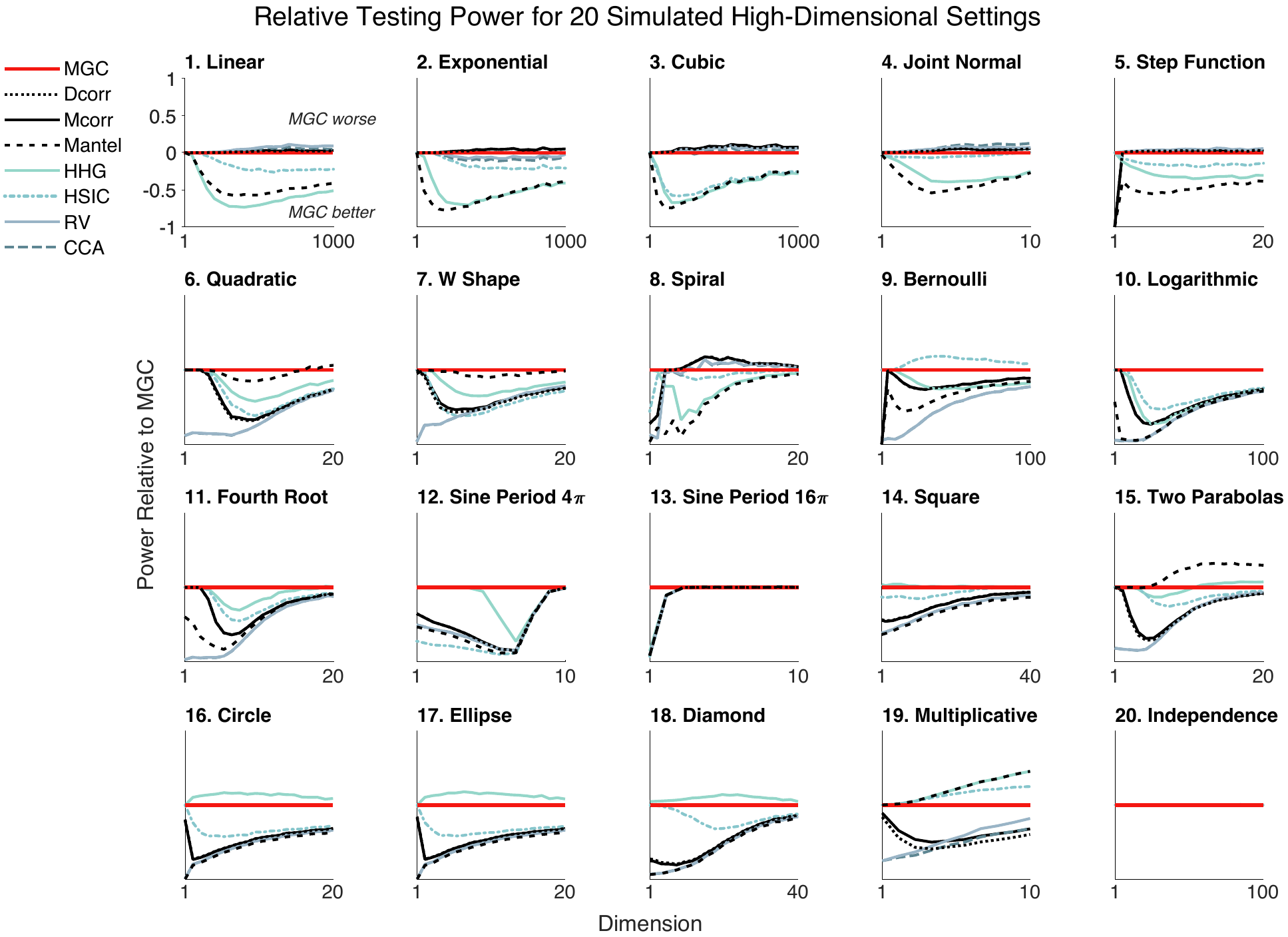}
\caption{An extensive benchmark suite of 20 different relationships spanning polynomial, trigonometric,  geometric, and other  relationships demonstrates that \Mgc~empirically nearly dominates eight other methods across dependencies and dimensionalities ranging from 1 to 1000.
% Power of different methods for $20$ different dependencies (see Algorithm \ref{alg:power} for details on computing power). It includes eight different tests: \Mgc~(solid  red), \Dcorr, \Mcorr, and \Mantel~(black dotted, solid, and dashed lines, respectively), \Hhg~and \Hsic~(solid and dash-dotted light green, respectively), \RV~and \CCA~(solid and dashed gray, respectively).
Each panel shows the testing power of other methods relative to the power of \Mgc~(e.g., power of \Mcorr~minus the power of \Mgc) at significance level $\alpha=0.05$ versus dimensionality for $n=100$. Any line below zero at any point indicates that that method's power is less than \Mgc's power for the specified setting and dimensionality. \Mgc~achieves empirically better  (or similar) power than all other methods in almost all relationships and all dimensions. For the the independent relationship (\#20), all methods yield power $0.05$ as they should. Note that \Mgc~is always plotted ``on top'' of the other methods, therefore, some lines are obscured.}
\label{f:nDAll}
\end{figure}

We generate an extensive benchmark suite of 20 relationships, including different polynomial (linear, quadratic, cubic), trigonometric (sinusoidal, circular, ellipsoidal, spiral), geometric (square, diamond, W-shape), and other functions.  This suite includes and extends the simulated settings from previous dependence testing work~\cite{SzekelyRizzoBakirov2007, SimonTibshirani2012, GorfineHellerHeller2012, HellerGorfine2013, SzekelyRizzo2013a}.
For many of them, we introduce high-dimensional variants, to more extensively evaluate the methods; 
function details are in Appendix~\ref{appen:function}.
 The visualization of one-dimensional noise-free (black) and noisy (gray) samples is shown in Supplementary Figure~\ref{f:dependencies}.
For each relationship, we compute the power of each method relative to  \Mgc~for  $\sim$20 different dimensionalities, ranging from 1 up to 10, 20, 40, 100, or 1000.
The high-dimensional relationships are more challenging because (1) they cannot be easily visualized, and (2) each dimension is designed to have less and less signal,	so there are many noisy dimensions.
Figure \ref{f:nDAll} shows that 
\Mgc~achieves the highest (or close to the highest)  power 
given 100 samples 
for each relationship and  dimensionality.
Supplementary Figure \ref{f:1DAll} shows the same advantage in one-dimension with increasing sample size. 

Moreover, for each relationship and each method we compute the required sample size to achieve power $85\%$ at error level $0.05$, and summarize the median size for monotone relationships (type 1-5) and non-monotone relationships (type 6-19) in Table~\ref{t:sim1}. 
Other methods typically require double or triple the number of samples as \Mgc~to achieve the same power. 
More specifically, traditional correlation methods (\Pearson, \RV, \CCA, \Spearman, \Kendall) always perform the best in monotonic simulations,  distance-based methods including \Mcorr, \Dcorr, \Mgc, \Hhg~and \Hsic~are slightly worse, while \Mic~and \Mantel~are the worst. \Mgc's performance is equal to linear methods on monotonic relationships.
For non-monotonic relationships, traditional correlations fail to detect the existence of dependencies, 
\Dcorr, \Mcorr, and \Mic, do reasonably well, but \Hhg~and \Mgc~require the fewest  samples.
In the high-dimensional non-monotonic relationships that motivated this work, and is common in biomedicine, \Mgc~significantly outperforms other methods.  In fact,  the second best method is \Mantel, and it requires $1.6\times$ as many  samples as \Mgc~to achieve the same power (in prior work we proved that \Mantel~is not a universally consistent test~\cite{mgc2}).  The second best test that is universally consistent (\Hhg) requires nearly double as many samples as \Mgc, demonstrating that \Mgc~could half the time and cost of experiments designed to discover relationships with a given effect size.

\Mgc~extends previously proposed global methods, such as \Mantel~and \Dcorr.  The above experiments extended \Mcorr, because \Mcorr~is universally consistent and an unbiased version of \Dcorr~\cite{SzekelyRizzo2013a}.  Supplementary Figure \ref{f:nDMantel} directly compares multiscale generalizations of \Mantel~and \Mcorr~as dimension increases, demonstrating that empirically, \Mgc~nearly dominates its global variant for essentially all dimensions and simulation settings considered here.   Supplementary Figure~\ref{f:1DMantel} shows a similar result for one-dimensional settings while varying sample size. Thus, not only does \Mgc~empirically nearly dominate existing tests, it is a framework that one can apply to future tests to further improve their performance.

%\begin{wraptable}{r}{0.6\linewidth}
\begin{table*}[!ht]
\centering
\caption{The median sample size for each method to achieve power $85\%$ at type 1 error level $0.05$, grouped into monotone (type 1-5) and non-monotone relationships (type 6-19) for both one- and ten-dimensional settings, normalized by the number of samples required by \Mgc. In other words, a $2.0$ indicates that the method requires double the sample size to achieve $85\%$ power relative to \Mgc.   \Pearson, \RV, and \CCA~all achieve the same performance, as do \Spearman~and \Kendall.
\Mgc~requires the fewest number of samples in all settings, and  for high-dimensional non-monotonic relationships, all other methods require about double or triple the number of samples   \Mgc~requires.}
\label{t:sim1}%
\begin{tabular}{l|l l l | l l l}
\toprule
Dimensionality&\multicolumn{3}{c}{One-Dimensional} & \multicolumn{3}{c}{Ten-Dimensional} \\
Dependency Type & Monotone & Non-Mono & Average & Monotone & Non-Mono&   Average \\
\midrule
 \Mgc  	& \textbf{1}  & \textbf{1} & \textbf{1} &\textbf{1} & \textbf{1} & \textbf{1} \\
  \Dcorr & \textbf{1}  & 2.6 & 2.2 &\textbf{1}  & 3.2 & 2.6\\
 \Mcorr	& \textbf{1}  & 2.8 & 2.4 &\textbf{1} & 3.1  & 2.6 \\
 \Hhg 	& 1.4  			 & \textbf{1} & 1.1 & 1.7  & 1.9 & 1.8  \\
\Hsic 		& 1.4  			& 1.1 & 1.2 & 1.7 & 2.4 & 2.2 \\
 \Mantel & 1.4  			& 1.8 & 1.7 & 3 & 1.6 & 1.9\\
\Pearson~/ \RV~/ \CCA & \textbf{1}  & >10 & >10 & \textbf{0.8} & >10 & >10 \\
\Spearman~/ \Kendall & \textbf{1}  			& >10 & >10 & n/a & n/a & n/a\\
\Mic & 2.4  																& 2 & 2.1 & n/a & n/a & n/a\\
%  Oracle \Mgc  & \textbf{50}  & 60 & \textbf{70} & \textbf{135} \\
% \Mgc  & \textbf{50}  & 60 & \textbf{90} & \textbf{165} \\
%  \Dcorr & \textbf{50}  & 60 & 235 & 535\\
% \Mcorr& \textbf{50}  & 60 & 250 & 515 \\
% \Hhg & 70  & 100 & \textbf{90} & 315  \\
%\Hsic & 70  & 100 & 95 & 400 \\
% \Mantel & 70  & 180 & 165 & 270\\
%\Pearson~/ \RV~/ \CCA & \textbf{50}  & \textbf{50} & >1000 & >1000 \\
%\Spearman~/ \Kendall & \textbf{50}  & n/a & >1000 & n/a \\
%\Mic & 120  & n/a & 180 & n/a\\
\bottomrule
\end{tabular}
\end{table*}
%\end{wraptable}
%}
% @EB  can  you make this  table pretty?

\subsection*{\Mgc~Decipher's Latent Dependence Structure}
\label{main3}

\begin{figure}[!ht]
\includegraphics[width=1.0\textwidth,trim={1cm 1cm 1cm 1cm},clip]{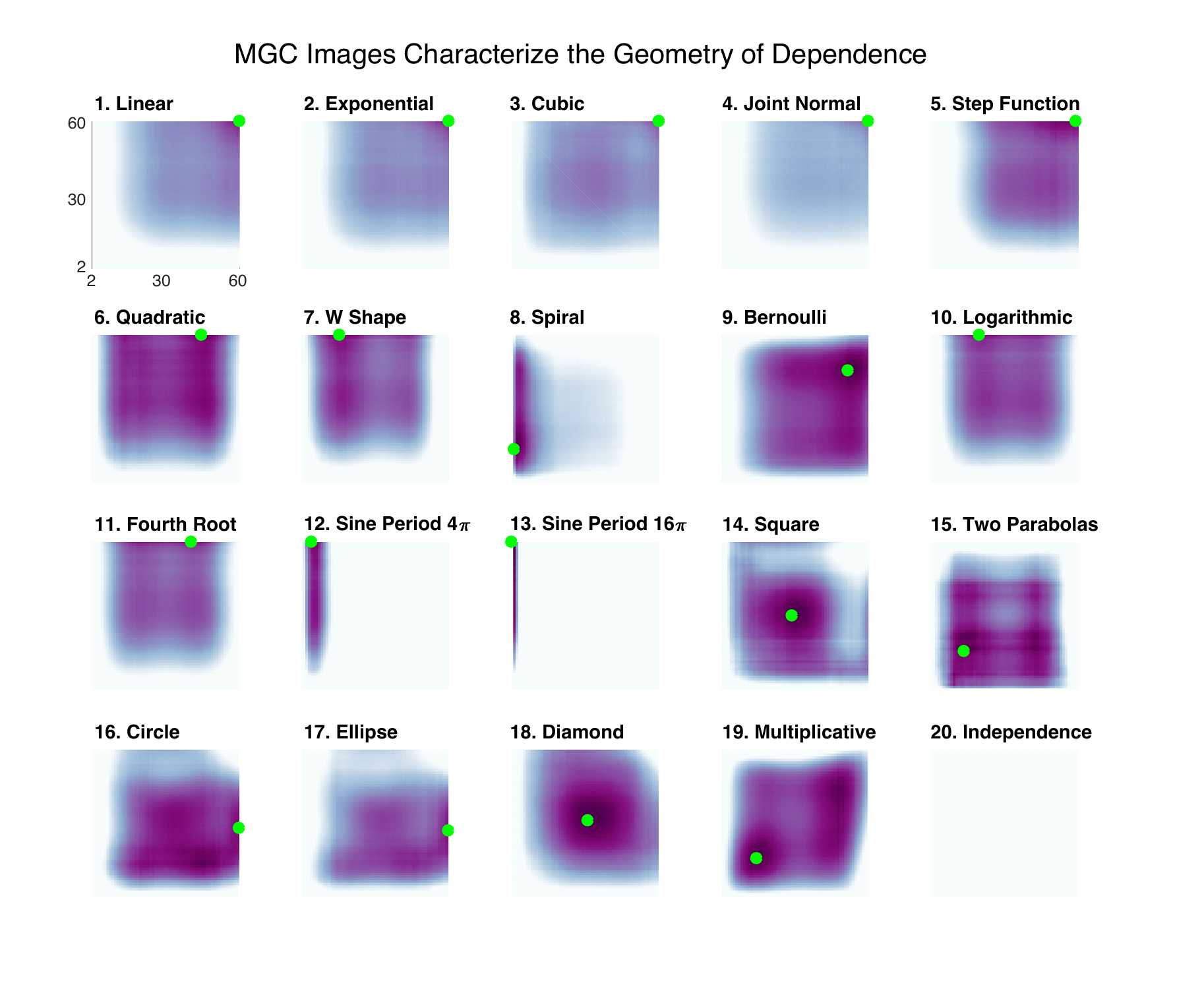}
\caption{The \Mgc-Map characterizes the geometry of the dependence function.
For each of the 20 panels, the abscissa and ordinate denote the number of neighbors for $X$ and  $Y$, respectively, and the color denotes the magnitude of each local correlation. For each simulation, the sample size is $60$, and both $X$ and $Y$ are one-dimensional. Each dependency has a different \Mgc-Map characterizing the geometry of dependence, and the optimal scale is shown in green.
In linear or close-to-linear relationships (first row), the optimal scale is global, i.e., the green dot is in the top right corner. Otherwise the optimal scale is non-global, which holds for the remaining dependencies. Moreover, similar dependencies often share similar \Mgc-Maps and similar optimal scales, such as (10) logarithmic and (11) fourth root, the trigonometric functions in (12) and (13), (16) circle and (17) ellipse, and (14) square and (18) diamond. A visualization of each dependency is provided in Appendix Figure~\ref{f:dependencies}, and the \Mgc-Maps for HD simulations are provided in Figure~\ref{f:powermaps1}.
}
\label{f:powermaps}
\end{figure}

Beyond simply discovering the existence of a relationship, the next goal is often to decipher the nature or structure of that relationship, thereby providing insight and guiding future experiments.
A single scalar quantity (such as effect size) is inadequate given the vastness and complexities of possible relationships.
Existing methods would require a secondary procedure to characterize the relationship, which introduces complicated ``post selection'' statistical quandaries that remain mostly unresolved~\cite{berk2013valid}.
Instead,  \Mgc~provides a simple, intuitive, and nonparametric (and therefore infinitely flexible) "map" of how it discovered the relationship. As described below, this map not only provides  interpretability for how \Mgc~detected a dependence, it also partially characterize the geometry of the investigated relationship.

The \emph{\Mgc-Map}  shows local correlation as a function of the scales of the two properties.  More concretely, it is the matrix of $t_{kl}$'s, as defined above.  Thus, the \Mgc-Map is an n-by-n matrix which encodes the strength of dependence for each possible scale.  
%, for a given dependence relationship, the local correlation as a function of the scales of the two properties.
Figure~\ref{f:powermaps} provides the \Mgc-Map for all 20 different one-dimensional relationships; the optimal scales, $\mh{t}_*$,  are shown with green dots.
For the monotonic dependencies (1-5), the optimal scale is always the largest scale, i.e., the global one.
For all non-monotonic dependencies (6-19),  \Mgc~chooses smaller scales.
Thus, a global optimal scale implies a close-to-linear dependency, otherwise the dependency is strongly nonlinear. In fact, this empirical observation led to the following theorem (which is proved in Appendix~\ref{appen:theory}) :
\begin{thm} \label{t:scale}
When $(\mbx,\mby)$ are linearly related (meaning that $Y$ can be constructed from $X$ by rotation, scaling, translation, and/or reflecti\Mgc-Mapon), the optimal scale of \Mgc~equals the global scale. Conversely, a local optimal scale implies a nonlinear relationship.
\end{thm}

Thus, the \Mgc-Map explains how \Mgc~discovers  relationships, specifically, which pairwise comparisons are most informative, and how that relates to the geometrical form of the relationship. 
%one can formally use \Mgc~not just to determine whether two properties are related, but also to determine aspects of the geometry of that relationship.  
Note that \Mgc~provides the geometric characterization ``for free'', meaning that no separate procedure is required; therefore, \Mgc~provides both a valid test and information about the geometric relationship. 
%We know of no other testing procedure that has this property.
%
Moreover, similar dependencies have similar \Mgc-Maps and often similar optimal scales. For example, logarithmic (10) and fourth root (11), though very different functions analytically, are geometrically similar, and yield very similar \Mgc-Maps.
Similarly,  (12) and (13) are trigonometric functions, and they share a narrow range of significant local generalized correlations.
Both circle (16) and ellipse (17), as well as square (14) and diamond (18), are closely related geometrically and also have similar \Mgc-Maps.
This indicates that the \Mgc-Map partially characterizes the geometry of these relationships, differentiating different dependence structures and assisting subsequent analysis steps.  Moreover, in~\cite{mgc2} we proved that the sample \Mgc-Map (which \Mgc~estimates) converges to the true \Mgc-Map provided by the underlying joint distribution of the data.
%
%we can prove the following theorem about the \Mgc-Map (see Appendix~\ref{appen:theory} for proof):
%\begin{thm}
%	The sample \Mgc-Map (which \Mgc~estimates) converges to the true \Mgc-Map provided by the underlying joint distribution of the data.
%\end{thm}
In other words, each relationship has a specific map that characterizes it based on its joint distribution, and \Mgc~is able to accurately estimate it via sample observations.  The existence of a population level characterization of the joint distribution strongly differentiates \Mgc~from previously proposed multi-scale geometric or topological characterizations of data, such as persistence diagrams~\cite{Edelsbrunner2009-zx}.

%\subsection*{\Mgc~Is Universally Consistent}
%\label{s:theory}
%
%\Mgc~generalizes any distance-based dependence test by computing a family of local correlations that iteratively exclude large distances per observation, then selects the smoothed maximum.
%%restricting it to only consider local distances.
%Any global test that \Mgc~generalizes is called \Mgc's ``global counterpart'', and we mainly consider \Mcorr~as the global counterpart to ensure consistency. The main theoretical result we obtain is as follows:
%%
%\begin{thm} \label{t:dominate}
%\Mgc~is valid and universally consistent.
%\end{thm}

%It is proved in \cite{mgc2} that the variance of \Mgc~and \Dcorr~are both $\mathcal{O}(1/n)$. Therefore, the advantage in magnitude is translated to the testing power, since they have the same rate of variance while \Mgc~can be much larger than \Mcorr~against nonlinear relationships; when the relationship is linear or close-to-linear, \Mgc~and \Mcorr~have almost the same test statistic and thus almost the same in testing power.

\paragraph*{\Mgc~is  Computationally Efficient}

\Mgc~is able to extend global methods without incurring large costs in computational time.
Though a na\"ive implementation of \Mgc~requires $\mc{O}(n^4)$ operations, we  devised a nested  implementation that requires only $\mc{O}(n^2 \log n)$ operations.
 Moreover, obtaining the \Mgc-Map  costs no additional computation, whereas other methods would require running a secondary computational step to decipher geometric properties of the relationship.
\Mgc, like existing methods, can also trivially be parallelized, reducing computation to  $\mc{O}(n^2 \log n / T)$, where $T$ is the number of cores  (see Algorithm \ref{alg:all_scales} for details).
 Since $T$ is often larger than $\log n$, in practice, \Mgc~can be $\mc{O}(n^2)$, meaning only a constant factor slower than its global counterpart.  
 For example, at sample size $n=5000$ and dimension $p=1$, on a typical laptop computer, \Mcorr~requires around $0.5$ seconds to compute the test statistic, whereas \Mgc~requires $5$ seconds. The cost and time to obtain $2.5 \times$ more data typically far exceeds a few seconds.  In comparison, the cost to compute a persistence diagram is typically $\mc{O}(n^3)$, which is orders of magnitude slower when $n >10$. The running time of each method on the real data experiments are reported in Appendix~\ref{appen:time}.

\subsection*{\Mgc~Uniquely Reveals  Relationships in Real Data}

Geometric intuition, numerical simulations, and theory all provide evidence that \Mgc~will be useful for real data discoveries.  Nonetheless, real data applications provide another necessary ingredient to justify its use in practice.  Below, we describe several real data applications where we have used \Mgc~to understand relationships in data that other methods were unable to provide.

\subsubsection*{\Mgc~Discovers the Relationships between Brain and Mental Properties}
%cs: Brain C*P, most similar to type 10 (log);
%Brain vs Disease, most similar to type 19 (multiplicative noise);
%Migrain vs CCI, most similar to type 1 (linear);
%Neuro-granin biomarker, most similar to type 5 or 9 (both are discrete);

% The study of the relationship between our physical and psychological properties dates back at least to the Greeks, if not further.  Neuroscientists, neurologists, psychiatrists, cognitive scientists, and even machine learning professionals seek to understand how the brain gives rise to the richness of the human experience.

%We investigate two particularly interesting properties of the human psyche: personality and creativity.  
%Although t
The human psyche is of course dependent on brain activity and structure.
Previous work has studied two particular aspects of our psyche: personality and creativity,
developing quantitative metrics for evaluating them using structured interviews \cite{Costa1992,Jung2009}.  
However, the relationship between brain activity and structure, and these aspects of our psyche, remains unclear.  We therefore utilized \Mgc~to published open access data to investigate.

%We use  previously published datasets to determine whether \Mgc~could yield insight into the relationship between our brains and these mental properties.

\begin{table*}[!ht]
\centering
\caption{The p-values for brain imaging vs mental properties. \Mgc~\emph{always} uncovers the existence of significant relationships and discovers the underlying optimal scales. Bold indicates significant p-value per dataset.}
\label{t:real}%
\begin{tabular}{|c||c|c|c|c|c|}
\hline
Testing Pairs / Methods & \Mgc & \Dcorr & \Mcorr & \Hhg & \Hsic \\
\hline
Activity vs Personality & $\textbf{0.043}$  & $0.667$ & $0.441$ & $0.059$ & $0.124$ \\
\hline
Connectivity vs Creativity & $\textbf{0.011}$  & $\textbf{0.010}$ & $\textbf{0.011}$ & $\textbf{0.031}$ & $0.092$ \\
\hline
\end{tabular}
\end{table*}

\begin{figure}
\begin{center}
\includegraphics[width=0.9\textwidth,trim={0cm 0cm 0cm 0cm},clip]{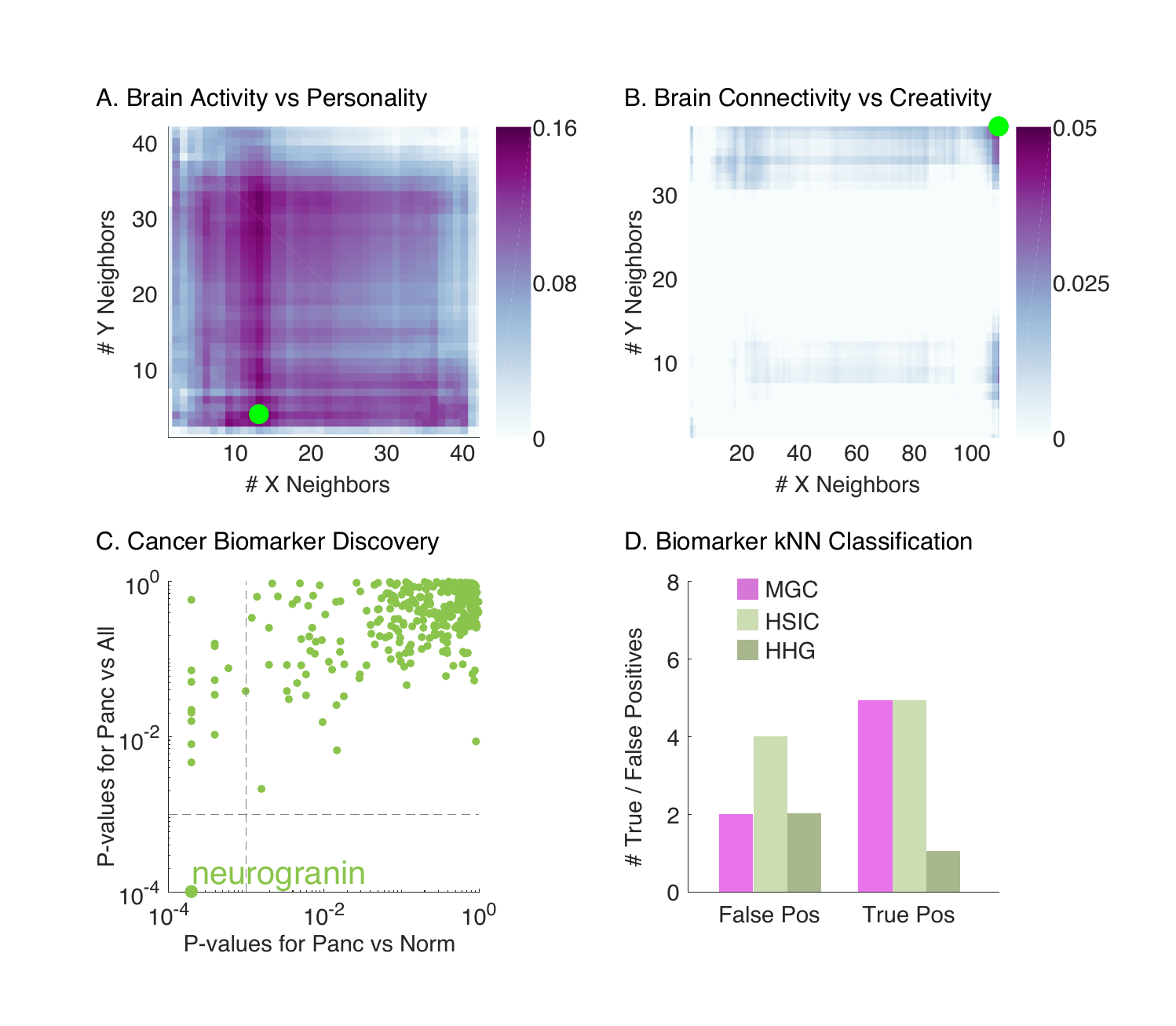}
\end{center}
\caption{Demonstration that \Mgc~successfully detects dependency, distinguishes linearity from nonlinearity, and identifies the most informative feature in a variety of real data experiments.
\textbf{(A)} The \Mgc-Map for brain activity versus personality. \Mgc~has a large test statistic and a significant p-value at the optimal scale $(13,4)$, while the global counterpart is non-significant. That the optimal scale is non-global implies a strongly nonlinear relationship.
\textbf{(B)} The \Mgc-Map for brain connectivity versus creativity. The image is similar to that of a linear relationship, and the optimal scale equals the global scale, thus both \Mgc~and \Mcorr~are significant in this case.
\textbf{(C)} For each peptide, the x-axis shows the p-value for testing dependence between pancreatic and healthy subjects by \Mgc, and the y-axis shows the p-value for testing dependence between pancreatic and all other subjects by \Mgc. At critical level $0.05$, \Mgc~identifies a unique protein after multiple testing adjustment.
\textbf{(D)} The true and false positive counts using a k-nearest neighbor  (choosing the best $k \in [1,10]$) leave-one-out classification using only the significant features identified by each testing method on the peptide data. The peptide identified by \Mgc~achieves the best true and false positive rates, as compared to the peptides identified by \Hsic~or \Hhg.
}
\label{f:real}
\end{figure}

First, we analyzed the relationship between resting-state functional magnetic resonance (rs-fMRI) activity and personality \cite{AdelsteinEtAl2011} (see Appendix \ref{app:personality} for details).
The first row of Table~\ref{t:real} compares the p-value of different methods, and Figure~\ref{f:real}{\color{magenta}A} shows the \Mgc-Map for the sample data. \Mgc~is able to yield a significant p-value ($< 0.05$), whereas all previously proposed global dependence tests under consideration (\Mantel, \Dcorr, \Mcorr, or \Hhg) fail to detect dependence at a significance level of $0.05$. Moreover, the \Mgc-Map provides a characterization of the dependence, for which the optimal scale indicates that the dependency is strongly nonlinear.
Interestingly, the \Mgc-Map does not look like any of the $20$ images from the simulated data, suggesting that the nonlinearity characterizing this dependency is more complex or otherwise different from those we have considered so far.

Second, we investigate the relationship between diffusion MRI derived connectivity and creativity \cite{Jung2009}  (see Appendix \ref{app:creativity} for details).
The second row of Table~\ref{t:real} shows that \Mgc~is able to ascertain a dependency between the whole brain network and the subject's creativity.
 The \Mgc-Map in Figure~\ref{f:real}{\color{magenta}B} closely resembles a linear relationship where the optimal scale is global. The close-to-linear relationship is also evident from the p-value table as all methods except \Hsic~are able to detect significant dependency, which suggests that there is relatively little to gain by pursuing nonlinear regression techniques, potentially saving valuable research time bu avoiding tackling an unnecessary problem. %Both \Mgc~and \Mcorr~equal $0.04$ in test statistic.
The test statistic for both \Mgc~and \Mcorr~equal $0.04$, which is quite close to zero despite a significant p-value, implying a relatively weak relationship.
A prediction of creativity via linear regression turns out to be non-significant, which implies that the sample size is too low to obtain useful predictive accuracy (not shown), indicating that more data are required for single subject predictions.
If one had first directly estimated the regression function, obtaining a null result, it would remain unclear whether  a relationship existed.  
This experiment demonstrates that for high-dimensional and potentially structured data, \Mgc~is able to reveal dependency with relatively small sample size while parametric techniques and  directly estimating regression functions can often be ineffective.

The performance in the real data closely matches the simulations in terms of the superiority of \Mgc: the first dataset is a strongly nonlinear relationship, for which \Mgc~has the lowest p-value, followed by \Hhg~and \Hsic~and then all other methods; the second dataset is a close-to-linear relationship, for which global methods often perform the best while \Hhg~and \Hsic~are trailing.  Moreover, \Mgc~detected a complex nonlinear relationship for brain activity versus personality, and a nearly linear relationship for brain network versus creativity, the only method able to make either of those claims. 
In a separate experiment, we assessed the frequency with which \Mgc~obtained false positive results using brain activity data, based on experiments from~\cite{EklundKnutsson2012,Eklund2015}.  Supplementary Figure~\ref{f:fpr} shows that \Mgc~achieves a false positive rate of $5\%$ when using a significance level of $0.05$, implying that it correctly controls for false positives, unlike typical parametric methods on these data.

\subsubsection*{\Mgc~Identifies Potential Cancer Proteomics Biomarkers}

\Mgc~can also be useful for a completely complementary set of scientific questions:
screening proteomics data for biomarkers, often involving the analysis of tens of thousands of proteins, peptides, or transcripts in multiple samples representing a variety of disease types. Determining whether there is a relationship between one or more of these markers and a particular disease state can be challenging, but is a necessary first step. We sought to discover new useful protein biomarkers from a quantitative proteomics technique that measures protein and peptide abundance called Selected Reaction Monitoring (SRM) \cite{PMID21248225} (see Appendix \ref{app:cancer} for details).
Specifically, we were interested in finding biomarkers that were unique to pancreatic cancer, because it is lethal and no clinically useful biomarkers are currently available.

The data consist of proteolytic peptides derived from the blood samples of  $95$ individuals harboring pancreatic ($n=10$), ovarian ($n=24$), colorectal cancer ($n=28$), and healthy controls  ($n=33$).
The processed data included $318$ peptides derived from $121$ proteins.
Previously, we used these data and other techniques to find ovarian cancer biomarkers (a much easier task because the dataset has twice as many ovarian patients) and validated them with subsequent experiments \cite{Wang2017}. Therefore, our first step was to check whether \Mgc~could correctly identify ovarian biomarkers. Indeed, the pepetides that have been validated previously are also identified by \Mgc~(see Appendix \ref{app:cancer}).
Emboldened, using the same dataset,
we applied \Mgc~to screen for biomarkers unique to pancreatic cancer.  To do so, we first screened  for a difference between pancreatic cancer and healthy controls, identifying several potential biomarkers.  Then, we screened for a difference between pancreatic cancer and all other conditions, to find peptides that differentiate pancreatic cancer from other cancers. Figure~\ref{f:real}{\color{magenta}C} shows the p-value of each peptide assigned by \Mgc, which  reveals one particular protein, neurogranin, that exhibits a strong dependency specifically with pancreatic cancer. Subsequent literature searches reveal that neurogranin is a potentially valuable biomarker for pancreatic cancer because it is exclusively expressed in brain tissue among normal tissues and has not been linked with any other cancer type. In comparison, \Hsic~identified neurogranin as well, but it also identified another peptide; \Hhg~identified the same two by \Hsic, and a third peptide.
A literature evaluation of these additional peptides shows that they are upregulated in other cancers as well and are unlikely to be useful as a pancreatic biomarker.
The rest of the global methods did not identify any markers.

We  carried out a classification task using the biomarkers identified by the various algorithms, using a  k-nearest-neighbor classifier to predict pancreatic cancer, and a leave-one-subject-out validation. Figure~\ref{f:real}{\color{magenta}D} shows that the peptide selected by \Mgc~(neurogranin) works better than any other subset of the peptides selected by \Hsic~or \Hhg, in terms of both fewer false positives and negatives.
%is the best peptide with $2$ false positives (among $85$ non-pancreatic) and $5$ true positives (among $10$ pancreatic), while the features identified by \Hsic~and \Hhg~yield higher false positives and lower true positives, either jointly or separately (excluding neurogranin). The classification result is shown in Figure~\ref{t:real}(iv),
This analysis suggests  \Mgc~can effectively be used for screening and subsequent classification.
%is working as intended and successfully discovers an ideal candidate for further experimentation.%

\subsection*{Discussion}
\label{conclu}

%We propose Multiscale Graph Correlation (\Mgc) to discover the presence and geometry of dependence across disparate types of data.
%We proved that Oracle \Mgc~dominates global approaches in finite samples.  Specifically, comparing the number of samples required to achieve a given power for a fixed significance value, \Mgc~requires the same number as  global approaches on linear relationships, but far fewer than global approaches for strongly nonlinear relationships. We further empirically demonstrate via simulations that \Mgc~nearly always outperforms (requires fewer samples) global methods regardless of the dimension, sample size, and geometry.  Moreover, \Mgc~provides a map indicating which scales are maximally informative about the dependence structure.
%In real data experiments, \Mgc~revealed dependency where global methods fail, as well as the geometry of those dependencies, and did not falsely detect signals when there were none.

There are a number of connections between \Mgc~and other prominent statistical procedures that may be worth further exploration. First,  \Mgc~can be thought of as a regularized or sparsified variant of  distance or kernel methods.  Regularization is central to high-dimensional and ill-posed problems, where dimensionality is larger than sample size. The connection made here between regularization and dependence testing opens the door towards considering other regularization techniques for correlation-based dependence testing, including \Hhg. %~and the \Mic~approach described in Reshef et al. \cite{Reshef2011}.
Second, \Mgc~can be thought of informally as learning a metric because it chooses amongst a set of $n^2$ truncated distances, motivating studying the relationship between \Mgc~and recent advances in metric learning \cite{xing2003distance}.  In particular, deep learning can be thought of as metric learning \cite{giryes2015deep}, and generative adversarial networks \cite{goodfellow2014generative} are implicitly testing for equality, which is closely related to dependence~\cite{Sutherland2016-fo}.
While \Mgc~searches over a two-dimensional parameter space to optimize the metric, deep learning searches over a much larger parameter space, sometimes including millions of dimensions.  Probably neither is optimal, and somewhere between the two would be useful in many tasks.
Third, energy statistics provide state of the art approaches to other problems, including goodness-of-fit \cite{Szekely2005}, analysis of variance \cite{Rizzo2010}, conditional dependence  \cite{Szekely2014,Wang2015}, and feature selection \cite{LiZhongZhu2012,Zhong2015}, so \Mgc~can be adapted for them as well.
In fact, \Mgc~can also implement a two-sample (or generally the K-sample) test \cite{Szekely2004, heller2016consistent, Shen2018-st}, so further comparisons of \Mgc~to standard methods for two-sample testing will be interesting.
Finally, although energy statistics have not yet been used for classification, regression, or dimensionality reduction, \Mgc~opens the door to these applications by providing guidance as to how to proceed.
Specifically, it is well documented in machine learning literature that the choice of kernel, metric, or scale often has an undesirably strong effect on the performance of different machine learning algorithms \cite{levina2004maximum}. \Mgc~provides a mechanism to estimate scale that is both theoretically justified and computationally efficient, by optimizing a metric for a task wherein the previous methods lacked a notion of optimization.  Nonlinear dimensionality reduction procedures, such as Isomap \cite{TenenbaumSilvaLangford2000} and local linear embedding \cite{SaulRoweis2000} for example, must also choose a scale, but have no principled criteria for doing so.  \Mgc~could  be used to provide insight into multimodal dimensionality reduction as well.

The fact that \Mgc~provides an estimate of the informative scales suggests several  theoretical steps to extend this work.
First, further theoretical guidance for choosing the optimal scale in \emph{finite} samples,  could possibly further improve performance.  Second, because the \Mgc-Maps provide insight into the geometry of dependence,  theoretically determining a mapping from these maps to the set of all nonlinear functions to provide a formal characterization of the geometry of the dependency, would be of interest.

\Mgc~also addresses a particularly vexing statistical problem that arises from the fact that methods  methods for discovering dependencies are typically dissociated from methods for deciphering them.
This dissociation creates a problem because the statistical assumptions underlying the ``deciphering'' methods become compromised in the process of  ``discoverying''; this is called the ``post-selection inference'' problem \cite{berk2013valid}.
The most straightforward way to address this issue is to collect new data, which is costly and time-consuming. Therefore, researchers often ignore this fact and make statistically invalid claims.
\Mgc~circumvents this dilemma by carefully constructing its permutation test to estimate the scale in the process of estimating a p-value, rather than after.
To our knowledge, \Mgc~is the first dependence test to take a step towards valid post-selection inference.

As a separate next theoretical extension, we could reduce the computational space and time required by \Mgc. \Mgc~currently requires space and time that are quadratic with respect to the number of samples, which can be costly for very large data.  Recent advances in related work suggest that we could reduce computational time to close to linear \cite{Huo2016}, although with some weakening of the theoretical guarantees \cite{zhang2017large}. Alternately, semi-external memory implementations would allow the running of \Mgc~on any data as long as the interpoint comparison matrix fits on disk rather than main memory \cite{Zheng2015,Zheng2016,Zheng2016c,Zheng2016b}. Another approach would be to derive an approximation to the asymptotic null distribution for \Mgc, obviating the need for the permutation test, but at the cost of potential finite-sample bias. %The fact that others have done so for kernel-based independence tests
%\cite{GrettonEtAl2005, GrettonGyorfi2010, GrettonEtAl2012}, which are equivalent to ``energy statistics'' (such as \Dcorr~and \Mcorr) \cite{SejdinovicEtAl2013, RamdasEtAl2015}, suggests that we could do so as well.

Finally, \Mgc~is easy to use: it merely requires pairs of samples to run, and all the code is available in both MATLAB  and as a R package on Comprehensive R Archive Network, available from \url{https://neurodata.io/tools/} and \url{https://github.com/neurodata/mgc} (code for reproducing all the figures in this manuscript is also available from the above websites)~\cite{Bridgeford2018-la}.  Because \Mgc~is open source and reproducible, and obtains near empirical dominance of other methods,
\Mgc~is situated to be useful in a wide range of applications.
We showed its value in diverse applications spanning neuroscience, which motivated this work, and an 'omics example.
Applications in other domains facing similar questions of dependence,
such as finance, pharmaceuticals, commerce, and security,  could likewise benefit from the methodology proposed here.

%\clearpage
%\pagestyle{empty}
\bibliographystyle{unsrtnat}
\bibliography{MGCbib}

\section*{Acknowledgment}
% \addcontentsline{toc}{section}{Acknowledgment}
This work was partially supported by
the Child Mind Institute Endeavor Scientist Program,
the National Science Foundation award DMS-1712947,
the National Security Science and Engineering Faculty Fellowship (NSSEFF),
the Johns Hopkins University Human Language Technology Center of Excellence (JHU HLT COE),
the Defense Advanced Research Projects Agency's (DARPA) SIMPLEX program through SPAWAR contract N66001-15-C-4041,
the XDATA program of DARPA administered through Air Force Research Laboratory contract FA8750-12-2-0303,
DARPA Lifelong Learning Machines program through contract FA8650-18-2-7834,
the Office of Naval Research contract N00014-12-1-0601,
the Air Force Office of Scientific Research contract FA9550-14-1-0033.
The authors thank Dr. Brett Mensh of Optimize Science for acting as our intellectual consigliere, Julia Kuhl for help with figures, and Dr.~Ruth Heller, Dr.~Bert Vogelstein, Dr.~Don Geman, and Dr.~Yakir Reshef for insightful suggestions.

\clearpage
\appendix
\setcounter{figure}{0}
\setcounter{thm}{0}
\renewcommand{\thealgorithm}{C\arabic{algorithm}}
\renewcommand{\thefigure}{E\arabic{figure}}
\renewcommand{\thesubsection}{\thesection.\Roman{subsection}}
\renewcommand{\thesubsubsection}{\thesubsection.\arabic{subsubsection}}

\section{Mathematical Details}
\label{appen:mgc}
This section contains essential mathematical details on independence testing, the notion of the generalized correlation coefficient and the distance-based correlation measure, how to compute the local correlations, and the smoothing technique. A more statistical treatment on \texttt{MGC} is in \cite{mgc2}, which introduces the population version of \Mgc~and various theoretical properties.

\subsection{Testing Independence}

Given pairs of observations $(\mb{x}_{i},\mb{y}_{i}) \in \Real^{p} \times \Real^{q}$ for $i=1,\ldots,n$, assume they are independently identically distributed as $(\mbx,\mby) \iid F_{\mbx \mby}$. If the two random variables \mbx~and \mby~are independent, the joint distribution equals the product of the marginals, i.e., $F_{\mbx \mby}=F_{\mbx} F_{\mby}$.  The statistical hypotheses for testing independence is as follows:
\begin{align*}
& H_{0}: F_{\mbx \mby}=F_{\mbx} F_{\mby},\\
& H_{A}: F_{\mbx \mby} \neq F_{\mbx} F_{\mby}.
\end{align*}
Given a test statistic, the testing power equals the probability of rejecting the independence hypothesis (i.e. the null hypothesis) when it is false. A test statistic is consistent if and only if the testing power increases to $1$ as sample size increases to infinity. We would like a test to be universally consistent, i.e., consistent against all joint distributions. \Dcorr, \Mcorr, \Hsic, and \Hhg~are all consistent against any joint distribution of finite second moments and finite dimension.

Note that $p$ is the dimension for $\mb{x}$'s, $q$ is the dimension for $\mb{y}$'s. For \Mgc~and all benchmark methods, there is no restriction on the dimensions, i.e., the dimensions can be arbitrarily large, and $p$ is not required to equal $q$. The ability to handle data of arbitrary dimension is crucial for modern big data. There also exist some special methods that only operate on one-dimensional data, such as \cite{Reshef2011,heller2016consistent,Huo2016}, which are not generalizable to multidimensional data. % and thus not further considered in this paper.

\subsection{Generalized Correlation}
Instead of computing on the sample observations directly, most state-of-the-art dependence tests operate on pairwise comparisons, either similarities (such as kernels) or dissimilarities (such as distances).

Let $\mathcal{X}_{n}=\{\mb{x}_{1},\cdots, \mb{x}_{n}\} \in \Real^{p \times n}$ and $\mathcal{Y}_{n}=\{\mb{y}_{1},\cdots, \mb{y}_{n}\} \in \Real^{q \times n}$ denote the matrices of sample observations, and $\delta_x$ be the distance function for $\mb{x}$'s and $\delta_y$ for $\mb{y}$'s.
One can then compute two $n \times n$ distance matrices $\tilde{A}=\{\tilde{a}_{ij}\}$ and $\tilde{B}=\{\tilde{b}_{ij}\}$, where $\tilde{a}_{ij}=\delta_x(\mb{x}_i,\mb{x}_j)$ and $\tilde{b}_{ij}=\delta_y(\mb{y}_i,\mb{y}_j)$. A common example of the distance function is the Euclidean metric ($L^{2}$ norm), which serves as the starting point for all methods in this manuscript. Note that we will use slightly different notations in the appendix: in the main paper $a_{ij}$ and $b_{ij}$ denote the Euclidean distance, while in the appendix they denote the centered distance with $\tilde{a}_{ij}$ and $\tilde{b}_{ij}$ denoting the Euclidean distance.

Let $A$ and $B$ be the transformed (e.g., centered) versions of the distance matrices $\tilde{A}$ and $\tilde{B}$, respectively. Any ``generalized correlation coefficient''  \cite{Spearman1904,KendallBook} can be written as:
\begin{equation}
\label{generalCoef}
\GG(\mathcal{X}_{n},\mathcal{Y}_{n})= \tfrac{1}{z} {\textstyle \sum_{i=1}^n \sum_{j=1}^n a_{ij} b_{ij}},
\end{equation}
where $z$ is proportional to the standard deviations of $A$ and $B$, that is $z=n^2\sigma_a \sigma_b$.
In words, $\GG$ is the global sample correlation across \emph{pairwise comparison matrices} $A$ and $B$, rather than the individual data samples.
A generalized correlation always has the range $[-1,1]$, has expectation $0$ under independence, and implies a stronger dependency when the correlation is further away from $0$.

%A generalized correlation coefficient therefore must make two choices. First, how to obtain the matrices $A$ and $B$.

Traditional correlations such as the Pearson's correlation and the rank correlation can be written as generalized correlation coefficients, where $A$ and $B$ are derived from sample observations rather than distances. Distance-based methods like \Dcorr~and \Mantel~operate on the distance metric, which may be chosen on the basis of domain knowledge, or by default they use the Euclidean distance; then transform the resulting distance matrices $\tilde{A}$ and $\tilde{B}$ by certain centering schemes into $A$ and $B$. \Hsic~chooses the Gaussian kernel and computes two kernel matrices, then transform the kernel matrices $\tilde{A}$ and $\tilde{B}$ by the same centering scheme as \Dcorr. For \Mgc, $A$ and $B$ are always distance matrices (or can be transformed to distances from kernels by \cite{SejdinovicEtAl2013}), and we shall apply a slightly different centering scheme that turns out to equal \Dcorr.

To carry out the hypothesis testing on sample data via a nonparametric test statistic, e.g., a generalized correlation, the permutation test is often an effective choice \cite{GoodPermutationBook}, because a p-value can be computed by comparing the correlation of the sample data to the correlation of the permuted sample data. The independence hypothesis is rejected if the p-value is lower than a pre-determined type $1$ error level, say $0.05$. Then the power of the test statistic equals the probability of a correct rejection at a specific type $1$ error level. Note that \Hhg~is the only exception that cannot be cast as a generalized correlation coefficient, but the permutation testing is similarly effective for the \Hhg~test statistic; also note that the \emph{iid} assumption is critical for permutation test to be valid, which may not be applicable in special cases like auto-correlated time series \cite{Mantel2013}.

%\subsubsection{The \Mantel~Coefficient}
%\label{appen:mantel}

%Define the overall mean of $\tilde{A}$ by $\bar{a}=\tfrac{1}{n^2}\sum_{i,j=1}^{n}(\tilde{a}_{ij})$ and similarly for $\tilde{B}$.
%The \Mantel~test defines
%\[a_{ij} = \left\{
 % \begin{array}{lr}
 %   \tilde{a}_{ij}-\bar{a}, & \mbox{ if } i \neq j, \\
 %   0, &\mbox{ if } i = j,
 % \end{array}
%\right.
%\]
%and similarly for $b_{ij}$.
%Unlike \Dcorr, \Mcorr, and \Hhg, the \Mantel~test does not yet have a consistency proof against all dependent alternatives,
%but it has been a very popular method in biology and ecology, possibly due to its simplicity and effectiveness. Figures~\ref{f:nDAll} and ~\ref{f:1DAll} indeed show that global \Mantel~is sub-optimal relative to much more recently proposed tests, and appears to be inconsistent for many dependencies.

\subsection{Distance Correlation (\Dcorr) and the Unbiased Version (\Mcorr)}
\label{appen:dcorr}

Define the row and column means of $\tilde{A}$ by $\bar{a}_{\cdot j}=\frac{1}{n} \sum_{i=1}^n \tilde{a}_{ij}$ and $\bar{a}_{i \cdot}=\frac{1}{n} \sum_{j=1}^n \tilde{a}_{ij} $.
\Dcorr~defines
\[a_{ij} = \left\{
  \begin{array}{lr}
    \tilde{a}_{ij}-\bar{a}_{i\cdot} - \bar{a}_{\cdot j} + \bar{a}, & \mbox{ if } i \neq j, \\
    0, &\mbox{ if } i = j,
  \end{array}
\right.
\]
and similarly for $b_{ij}$.
For distance correlation, the numerator of Equation~\ref{generalCoef} is named the distance covariance (\Dcov), while $\sigma_a$ and $\sigma_b$ in the denominator are named the distance variances. The centering scheme is important to guarantee the universal consistency of \Dcorr, whereas \Mantel~uses a simple centering scheme and thus not universal consistent.

Let $c(\mbx,\mby)$ be the population distance correlation, that is, the distance correlation between the underlying random variables $\mbx$ and $\mby$. Szekely et al. (2007) define the population distance correlation via the characteristic functions of $F_{\mbx}$ and $F_{\mby}$, and show that the population distance correlation equals zero if and only if $\mbx$ and $\mby$ are independent, for any joint distribution $F_{\mbx \mby}$ of finite second moments and finite dimensionality.
They also show that  as $n \rightarrow \infty$, the sample distance correlation converges to the population distance correlation, that is, $\GG(\mathcal{X}_{n},\mathcal{Y}_{n}) \rightarrow c(\mbx,\mby)$. Thus the sample distance correlation is consistent against any dependency of finite second moments and dimensionality.
Of note, the distance covariance, distance variance, and distance correlation are always non-negative.  Moreover,  the consistency result holds for a much larger family of metrics, those of strong negative type  \cite{Lyons2013}.
%Note that the \Dcorr~here equals the square of distance correlation in \cite{SzekelyRizzoBakirov2007}, but for ease of presentation the square naming is dropped here.

It turns out that the sample distance correlation has a finite-sample bias, especially as the dimension $p$ or $q$ increases \cite{SzekelyRizzo2013a}. For example, for independent Gaussian distributions, the sample distance correlation converges to $1$ as $p, q \rightarrow \infty$.
%which not only makes the interpretation of distance correlation more difficult, but also impairs the testing power of \Dcorr~for high-dimensional data with finite sample size.
By excluding the diagonal entries and slightly modifies the off-diagonal entries of $\mathcal{A}$ and $\mathcal{B}$, Szekely and Rizzo (2013) \cite{SzekelyRizzo2013a, SzekelyRizzo2014, RizzoSzekely2016} show that
\Mcorr~is an unbiased estimator of the population distance correlation $c(\mb{x},\mb{y})$ for all $p, q, n$, which is approximately normal even if $p,q \rightarrow \infty$. Thus it enjoys the same theoretical consistency as \Dcorr~and always has zero mean under independence, which is the default choice \Mgc~is based on in this paper.  %Note that the \Mcorr~here is slightly different from the \Mcorr~in \cite{SzekelyRizzo2013a} because we define the diagonals of $A$ and $B$ differently, but the test statistic has only negligible difference and almost always the same testing performance.

\subsection{Local Generalized Correlations}
\label{appen:localCorr}

Local generalized correlations can be thought of as further generalizations of generalized correlation coefficients. In particular, given any matrices $A$ and $B$, we can define a set of local variants of them as follows. Let $R(A_{\cdot j},i)$ be the ``rank'' of $\mb{x}_i$ relative to $\mb{x}_j$, that is, $R(A_{\cdot j},i)=k$ if $\mb{x}_i$ is the $k^{th}$ closest point (or ``neighbor'') to $\mb{x}_j$, as determined by ranking the $n-1$ distances to $x_j$.
Define $R(B_{i \cdot},j)$ equivalently for the \mby's, but ranking relative to the rows rather than the columns (see below for explanation).
For any neighborhood size $k$ around each $\mb{x}_i$~and any neighborhood size $l$ around each $\mb{y}_j$, we define the local pairwise comparisons:
\begin{equation}
\label{localCoef2}
    \mt{a}_{ij}^k=
    \begin{cases}
      a_{ij}, & \text{if } R(A_{\cdot j},i) \leq k, \\
      0, & \text{otherwise};
    \end{cases} \qquad \qquad
    \mt{b}_{ij}^l=
    \begin{cases}
      b_{ij}, & \text{if } R(B_{i \cdot},j) \leq l, \\
      0, & \text{otherwise};
    \end{cases}
\end{equation}
and then let $a^k_{ij}=\mt{a}^k_{ij} - \bar{a}^k$,
where $\bar{a}^k$ is the mean of $\{\mt{a}_{ij}^{k}\}$, and similarly for $b^l_{ij}$.

The \emph{local} variant of any global generalized correlation coefficient is defined to effectively excludes large distances:
\begin{equation}
\label{localCoef}
\GG^{kl}(\mathcal{X}_{n},\mathcal{Y}_{n})=\dfrac{1}{z_{kl}} {\textstyle \sum_{i,j=1}^n a_{ij}^k b_{ij}^l},
\end{equation}
where $z_{kl}=n^2 \sigma_a^k \sigma_b^l$,  with $\sigma_a^k$ and $\sigma_b^{l}$ is the standard deviations for the truncated pairwise comparisons. Thus, $c^{kl}$ is the local sample generalized correlation at a given scale. The \Mgc-Map can be constructed by computing all local generalized correlations, which allows the discovery of the optimal correlation. For any aforementioned generalized correlation (\Dcorr, \Mcorr, \Hsic, \Mantel, \Pearson), its local generalized correlations can be directly defined by Equation~\ref{localCoef}, by plugging in the respective $a_{ij}$ and $b_{ij}$ from Equation~\ref{generalCoef}.

\subsection{\Mgc~as the Optimal Local Correlation}
\label{appen:mgc2}

We define the multiscale graph correlation statistic as the optimal local correlation, for which the family of local correlation is computed based on Euclidean distance and \Mcorr~transformation.

Instead of taking a direct maximum, \Mgc~takes a smoothed maximum, i.e., the maximum local correlation of the largest connected component $R$ such that all local correlations within $R$ are significant.
If no such region exists, \Mgc~defaults the test statistic to the global correlation (details in Algorithm \ref{alg:sample_mgc}).
Thus, we can write:
\begin{align} \label{eq:sampc}
&\GG^{*}(\mathcal{X}_{n},\mathcal{Y}_{n}) =  \max_{(k,l) \in R}c^{kl}(\mathcal{X}_{n},\mathcal{Y}_{n}) \\
&R=\mbox{Largest Connected Component of } \{(k,l) \mbox{ such that } c^{kl} > \max(\tau, c^{nn})\}. \nonumber
\end{align}
Then the optimal scale equals all scales within $R$ whose local correlations are as large as $\GG^{*}$. The choice of $\tau$ is made explicit in the pseudo-code, with further discussion and justification offered in \cite{mgc2}.

\subsection{Proof for Theorem~\ref{thm:main}}
\label{appen:theory}

\begin{thm}
\label{thm:main}
When $(\mbx,\mby)$ are linearly related (rotation, scaling, translation, reflection), the optimal scale of \Mgc~equals the global scale. Conversely, that. the optimal scale is local implies a nonlinear relationship.
\end{thm}
\begin{proof}

It suffices to prove the first statement, then the second statement follows by contrapositive.
When $(\mbx,\mby)$ are linearly related, $\mby=W\mbx+b$ for a unitary matrix $W$ and a constant $b$ up-to possible scaling, in which case the distances are preserved, i.e., $\|y_i - y_j\| = \|W x_i - W x_j\|=\|x_i - x_j\|$. It follows that $\Mcorr(\mathcal{X}_{n},\mathcal{Y}_{n})=1$, so the global scale achieves the maximum possible correlation, and the largest connected region $R$ is empty. Thus the optimal scale is global and $\Mgc(\mathcal{X}_{n},\mathcal{Y}_{n})=\Mcorr(\mathcal{X}_{n},\mathcal{Y}_{n})=1$.
\end{proof}

\subsection{Computational Complexity}
\label{appen:comp_comp}

The distance computation takes $\mc{O}(n^2 \max\{p,q\})$, and the ranking process takes $\mc{O}(n^2 \log n)$. Once the distance and ranking are completed, computing one local generalized correlation requires $\mc{O}(n^2)$ (see Algorithm \ref{alg:1scale}). Thus a naive approach to compute all local generalized correlations requires at least $\mc{O}(n^2 \max\{n^2, p,q\})$ by going through all possible scales, meaning possibly $\mc{O}(n^4)$ which would be computationally prohibitive. However, given the distance and ranking information, we devised an algorithm that iteratively computes all local correlations in $\mc{O}(n^2)$ by re-using adjacent smaller local generalized correlations (see Algorithm \ref{alg:all_scales}).
Therefore, when including the distance computation and ranking overheads, the MGC statistic is computed in $\mc{O}(n^2 \max\{\log n,p,q\})$), which has the same running time as the \Hhg~statistic, and the same running time up to a factor of $\log n$ as  global correlations like \Dcorr~and \Mcorr, which require  $\mc{O}(n^2 \max\{p,q\})$ time.

By utilizing a multi-core architecture, \Mgc~can be computed in $\mc{O}(n^2 \max\{\log n,p,q\}/T)$ instead. As $T=\log(n)$ is often a small number, e.g., $T$ is no more than $30$ at $1$ billion samples, thus \Mgc~can be effectively computed in the same complexity as \Dcorr. Note that the permutation test adds another $r$ random permutations to the $n^2$ term, so computing the p-value requires $\mc{O}(n^2 \max\{\log n,p,q, r\}/T)$.

\clearpage

\section{\Mgc~Algorithms and Testing Procedures}
\label{appen:algorithms}

Six algorithms are presented in order:
\begin{description}
\item[1.] Algorithm~\ref{alg:mgc} describes \Mgc~in its entirety (which calls most of the other algorithms as functions).
\item[2.] Algorithm~\ref{alg:power} evaluates the testing power of \Mgc~by a given distribution.
\item[3.] Algorithm~\ref{alg:sample_mgc} computes the \Mgc~test statistic.
\item[4.] Algorithm~\ref{alg:pval} computes the p-value of \Mgc~by the permutation test.
\item[5.] Algorithm~\ref{alg:1scale} computes the local generalized correlation coefficient at a given scale $(k,l)$, for a given choice of the global correlation coefficient.
\item[6.] Algorithm~\ref{alg:all_scales} efficiently computes all local generalized correlations, in nearly the same running time complexity as computing one local generalized correlation.
\end{description}
For ease of presentation, we assume there are no repeating observations of \mbx~or \mby, and note that \Mcorr~is the global correlation choice that \Mgc~builds on.

\clearpage
\begin{algorithm}
\caption{Multiscale Graph Correlation (\Mgc);  requires  $\mc{O}(n^2 \max(\log{n}, p,q, r)/T)$ time, where $r$ is the number of permutations and $T$ is the number of cores available for parallelization.}
\label{alg:mgc}
\begin{algorithmic}%[1]
\Require $n$ samples of $(x_i,y_i)$ pairs, an integer $r$ for the number of random permutations.
\Ensure (i) MGC statistic $\GG^*$, (ii) the optimal scale $(k,l)$,
(iii) the p-value $p(\GG^*)$,
%(iv) the multiscale correlation $\mathcal{C}$ and significance $\mathcal{P}$ maps, and (v) the estimated optimal scales $\widehat{\mathcal{KL}}^{*}$.
\Function{MGC}{$(x_i,y_i)$, for $i \in [n]$}
\Statex{\textbf{(1)} Calculate all pairwise distances:}
\For{$i,j:=1,\ldots,n$}
\State  $a_{ij} = \delta_x(x_i,x_j)$
\Comment{$\delta_x$ is the distance between pairs of $x$ samples}
\State  $b_{ij} = \delta_y(y_i,y_j)$
\Comment{$\delta_y$ is the distance between pairs of $y$ samples}
\EndFor
\State Let $A=\{ a_{ij}\}$ and $B=\{ b_{ij}\}$.
\Statex{\textbf{(2)} Calculate Multiscale Correlation Map $\mathcal{C}$ \& \Mgc~Test Statistic:}
%\State  $\mathcal{C}=\textsc{MGCAllLocal}(A,B)$  \Comment{local  correlation for all scales using Algorithm \ref{alg:all_scales}}
\State  $[\GG^*,\mathcal{C},k,l]=\textsc{MGCSampleStat}(A,B)$
\Comment{Algorithm \ref{alg:sample_mgc}}
\Statex{\textbf{(3)} Calculate the p-value }%$p(\hat{\GG}^*)$ and the set of estimated optimal scales $\widehat{\mathcal{KL}}^{*}$ from Sample \Mgc, as well as the multiscale significance map $\mathcal{P}$:}
\State $pval(\GG^*)=\textsc{PermutationTest}(A,B,r,\GG^*)$
\Comment{Algorithm~\ref{alg:pval}}
\EndFunction
\end{algorithmic}
\end{algorithm}

\clearpage

\begin{algorithm}
\caption{Power computation of \Mgc~against a given distribution.
%This algorithm computes the power for both Sample and Oracle \Mgc, as well as the multiscale power map (i.e., testing powers of all local generalized correlations).
By repeatedly sampling from the joint distribution $F_{\mbx \mby}$, sample data of size $n$ under the null and the alternative are generated for $r$ Monte-Carlo replicates. %Then all local generalized correlations under the null and the alternative hypotheses are computed by Algorithm~\ref{alg:all_scales}.
The power of \Mgc~follows by computing the test statistic under the null and the alternative using Algorithm~\ref{alg:sample_mgc}.
%Oracle \Mgc~directly maximizes the power map, obtainable by computing the testing power at each local generalized correlation.
%The running time is $\mc{O}(rn^2 \log n/T)$.
In the simulations we use $r=10$,$000$ MC replicates.
%This algorithm can be similarly adapted to training data, for which the alternative statistic can be computed from the training data while the null statistic can be computed by permutation.
Note that power computation for other benchmarks follows from the same algorithm by plugging in the respective test statistic. }
\label{alg:power}
\begin{algorithmic}[1]
\Require A joint distribution $F_{\mbx \mby}$, the sample size $n$, the number of MC replicates $r$, and the type $1$ error level $\alpha$.
\Ensure The power $\beta$ of \Mgc.
%the power of Oracle \Mgc~$\beta(\GG^{*})$, the Oracle power map $\{\beta_{kl}\} \in [0,1]^{n \times n}$, and the set of true optimal scales $\mathcal{KL}^{*}$.
\Function{MGCPower}{$F_{\mbx \mby}$, $n$, $r$, $\alpha$}
\For{$t:=1,\ldots,r$}
\For{$i:=[n]$}
\State $x^{0}_{i} \stackrel{iid}{\sim} F_{\mbx}$, $y^{0}_{i} \stackrel{iid}{\sim} F_{\mby}$  \Comment{sample from null}
\State	$(x^{1}_{i},y^{1}_{i}) \stackrel{iid}{\sim} F_{\mbx \mby}$, \Comment{sample from alternative}
\EndFor
\For{$i,j:=1,\ldots,n$}
\State $a^{0}_{ij} = \delta_x(x^{0}_i,x^{0}_j)$, $b^{0}_{ij} = \delta_y(y^{0}_i,y^{0}_j)$ \Comment{pairwise distances under the null}
\State $a^{1}_{ij} = \delta_x(x^{1}_i,x^{1}_j)$, $b^{1}_{ij} = \delta_y(y^{1}_i,y^{1}_j)$ \Comment{pairwise distances under the alternative}
\EndFor
%\State $\mathcal{C}_{0}[t]=\textsc{MGCAllLocal}(A^{0},B^{0})$ \Comment{all local generalized correlations under  null}
%\State $\mathcal{C}_{1}[t]=\textsc{MGCAllLocal}(A^{1},B^{1})$ \Comment{all local generalized correlations under alternative}
%
\State $\GG^{*}_{0}[t]=\textsc{MGCSampleStat}(A^{0},B^{0})$ \Comment{\Mgc~statistic under the null}
\State $\GG^{*}_{1}[t]=\textsc{MGCSampleStat}(A^{1},B^{1})$ \Comment{\Mgc~statistic under the alternative}
\EndFor

%\For{$k,l:=1,\ldots,n$} \Comment{for each scale}
%\State $\omega_{\alpha} \rto \textsc{Cdf}_{1-\alpha}(\GG_{0}^{kl}[t],t \in [r])$ \Comment{get the critical value from the empirical distributions}
%\State $\beta_{kl} \rto \sum_{t=1}^{r}(\GG_{1}^{kl}[t]>\omega_{\alpha}) / r$ \Comment{compute  power for each scale}
%\EndFor
%\State $\beta(\GG^{*}) \rto \max_{kl} \{\beta_{kl}\}$  \Comment{testing power of Oracle \Mgc}
%\State $\mathcal{KL}^{*} \rto \{(k,l)=\argmax_{\{k,l\}}\beta_{kl}\}$ \Comment{the set of scales that maximize the power}
\State $\omega_{\alpha} \rto \textsc{Cdf}_{1-\alpha}(\GG_{0}^{*}[t],t \in [r])$ \Comment{the critical value of \Mgc~under the null}
\State $\beta \rto \sum_{t=1}^r(\GG_{1}^{*}[t]>\omega_{\alpha}) / r$  \Comment{compute power by the alternative distribution}
\EndFunction
\end{algorithmic}
\end{algorithm}

\begin{algorithm}
\caption{\Mgc~test statistic. This algorithm computes all local correlations, take the smoothed maximum, and reports the $(k,l)$ pair that achieves it. For the smoothing step, it: (i) finds the largest connected region in the correlation map, such that each correlation is significant, i.e., larger than a certain threshold to avoid correlation inflation by sample noise, (ii) take the largest correlation in the region, (iii) if the region area is too small, or the smoothed maximum is no larger than the global correlation, the global correlation is used instead. The running time is $\mc{O}(n^2)$.}
\label{alg:sample_mgc}
\begin{algorithmic}[1]
\Require A pair of distance matrices $(A, B) \in \Real^{n \times n} \times \Real^{n \times n}$.
\Ensure The \Mgc~statistic $\GG^{*} \in \Real$, all local statistics $\mathcal{C} \in \Real^{n \times n}$, and the corresponding local scale $(k,l) \in \mathbb{N} \times \mathbb{N}$.
\Function{MGCSampleStat}{$A,B$}
\State $\mathcal{C}=\textsc{MGCAllLocal}(A,B)$ \Comment{All local correlations}
\State $\tau = \textsc{Thresholding}(\mathcal{C})$ \Comment{find a threshold to determine large local correlations}
\Linefor{$i,j := 1,\ldots, n$}{$r_{ij} \rto \mathbb{I} (c^{ij} > \tau )$} \Comment{identify all scales with large correlation}
\State $\mathcal{R} \rto \{r_{ij} : i,j = 1,\ldots, n\}$ \Comment{binary map encoding  scales with large correlation}
\State $\mathcal{R}  = \textsc{Connected}(\mathcal{R} )$ \Comment{largest connected component of the binary matrix}
\State $\GG^{*} \rto \mathcal{C}(n,n)$ \Comment{use the global correlation by default}
\State $k \rto n, l \rto n$
\If{$\left(\sum_{i,j} r_{ij}\right) \geq 2n $} \Comment{proceed when the significant region is sufficiently large}
\State $[\GG^{*},k,l] \rto \max (\mathcal{C} \circ \mathcal{R})$ \Comment{find the smoothed maximum and the respective scale}
%\State $\Omega \rto \{(i,j) :  \GG^{ij}\geq \max (\mathcal{C} \circ \mathcal{R})\}$ \Comment{scales with largest correlation in $\mathcal{R}$}
%\For{$(k',l') \in \Omega$}
%\State $\eta \rto \min_{k \in [k'-\gamma,k'+\gamma]}\{\GG^{kl'}\}$ \Comment{minimal corr on a fixed column}
%\State $k \rto \arg\min_{k \in [k'-\gamma,k'+\gamma]}\{\GG^{kl'}\}$ \Comment{the respective row index}
%\Lineif{$\eta \geq \hat{\GG}^{*}$}{ $\hat{\GG}^{*} \rto \eta, \hat{k} \rto k, \hat{l} \rto l'$}
%\State $\eta \rto \min_{l \in [l'-\gamma,l'+\gamma]}\{\GG^{k'l}\}$ \Comment{minimal corr a fixed row}
%\State $l \rto \arg\min_{l \in [l'-\gamma,l'+\gamma]}\{\GG^{k'l}\}$ \Comment{the respective column index}
%\Lineif{$\eta \geq \hat{\GG}^{*}$}{ $\hat{\GG}^{*} \rto \eta, \hat{k} \rto k', \hat{l} \rto l$}
%\EndFor
\EndIf
\EndFunction
\Statex
\Require $\mathcal{C} \in \Real^{n \times n}$.
\Ensure A threshold $\tau$ to identify large correlations.
\Function{Thresholding}{$\mathcal{C}$}
\State $\tau \rto \sum_{\GG^{ij}<0} (\GG^{ij})^2 / \sum_{\GG^{ij}<0} 1$ \Comment{variance of all negative local generalized correlations}
\State $\tau \rto \max\{0.01,\sqrt{\tau}\} \times 3.5$ \Comment{threshold based on negative correlations}
% \State $\tau_{2} \rto 2/n$
\State $\tau \rto \max\{\tau,2/n, \GG^{nn}\}$
\EndFunction
\end{algorithmic}
\end{algorithm}

\clearpage

\begin{algorithm}
\caption{Permutation Test.
This algorithm uses the random permutation test with $r$ random permutations for the p-value, %resulting in the p-value, the estimated optimal scales, and the multiscale significance map,
requiring $\mc{O}(rn^2 \log n)$ for \Mgc. %Specifically, it computes the p-values by comparing the multiscale correlation map and the sample \Mgc~statistic of the observed data, to those of each permuted resample.  Then, the optimal scales are estimated by taking the largest rectangle with local statistics no smaller than Sample \Mgc~and local significance values no larger than the p-value.
In the real data experiment we always set $r=10$,$000$. Note that the p-value computation for any other global generalized correlation coefficient follows from the same algorithm by replacing \Mgc~with the respective test statistic.
}
\label{alg:pval}
\begin{algorithmic}[1]
\Require A pair of distance matrices $(A, B) \in \Real^{n \times n} \times \Real^{n \times n}$, the number of permutations $r$, %the local generalized correlation map $\mathcal{C}$
and \Mgc~statistic $\GG^*$ for the observed data.
\Ensure The p-value $pval \in [0,1]$.% for Sample \Mgc, the set of estimated optimal scale $\widehat{\mathcal{KL}}^{*}$, and the p-value matrix $\mathcal{P} \in [0,1]^{n \times n}$ of all local generalized correlations.
\Function{PermutationTest}{$A$, $B$, $r$, $\GG^*$}
\For{$t:=1,\ldots,r$}
\State $\pi=\textsc{RandPerm}(n)$ \Comment{generate a random permutation of size $n$}
\State $\GG^{*}_{0}[t]=\textsc{MGCSampleStat}(A, B(\pi,\pi))$ \Comment{calculate the permuted \Mgc~statistic}
%\State $\hat{\GG}^{*}_{0}[t]=\textsc{SampleMGC}(\mathcal{C}_{0}[t])$ \Comment{calculate the permuted Sample \Mgc}
\EndFor

%\Linefor{$k,l:=1,\ldots,n$}{$p_{kl} \rto \sum_{t=1}^{r}(\GG^{kl} \leq \GG^{kl}_{0}[t])/r$}  $\mathcal{P} \rto \{ p_{kl} \}$
% \Comment{the significance map}
\State $pval(\GG^*) \rto \frac{1}{t}\sum_{t=1}^{r}\mb{I}(\GG^{*} \leq \GG^{*}_{0}[t])$  \Comment{compute p-value of \Mgc}
%\State Construct the binary map:    $\mathcal{E}^{kl} = 1 $ iff $  c^{kl} \geq \hat{\GG}^{*} \mbox{ and } p_{kl} \leq p(\hat{\GG}^*)$.
%\State $\widehat{\mathcal{KL}}^{*} \rto $ the set of elements in the largest axis-aligned rectangle in $\mathcal{E}$ containing only $1$'s.

\EndFunction
\end{algorithmic}
\end{algorithm}

\clearpage

\begin{algorithm}
\caption{Compute local test statistic at a given scale. This algorithm runs in $\mc{O}(n^2)$ once the rank information is provided, which is suitable for \Mgc~computation if an optimal scale is already estimated. But it would take $\mc{O}(n^4)$ if used to compute all local generalized correlations. Note that for the default \Mgc~implementation uses single centering, the centering function centers $A$ by column and $B$ by row, and the sorting function sorts $A$ within column and $B$ within row. By utilizing $T=\log(n)$ cores, the sorting function can be easily parallelized to take $\mc{O}(n^2 \log(n)/T)=\mc{O}(n^2)$.}
\label{alg:1scale}
\begin{algorithmic}[1]
\Require A pair of distance matrices $(A, B) \in \Real^{n \times n} \times \Real^{n \times n}$, and a  local scale $(k, l) \in \mathbb{N} \times \mathbb{N}$.
\Ensure The local generalized correlation coefficient $\GG^{kl} \in [-1,1]$.
\Function{LocalGenCorr}{$A$, $B$, $k$, $l$}
\Linefor{$Z:=A,B$}{$\mathcal{E}^{Z}=\textsc{Sort}(Z)$} \Comment{parallelized sorting}
\Linefor{$Z:=A,B$}{$Z=\textsc{Center}(Z)$}  \Comment{center distance matrices}
\State $\tilde{\GG}^{kl} \rto tr((A \circ \mathcal{E}^{A})\TT \times (B \circ (\mathcal{E}^{B})\TT))$ \Comment{un-normalized local distance covariance}
\State $v^{A} \rto tr((A \circ \mathcal{E}^{A})\TT \times (A \circ (\mathcal{E}^{A})\TT))$ \Comment{local distance variances}
\State $v^{B} \rto tr((B \circ \mathcal{E}^{B})\TT \times (B \circ (\mathcal{E}^{B})\TT))$
\State $e^{A} \rto \sum_{i,j=1}^{n}(A \circ \mathcal{E}^{A})_{ij}$ \Comment{sample means}
\State $e^{B} \rto \sum_{i,j=1}^{n}(B \circ \mathcal{E}^{B})_{ij}$
\State $\GG^{kl} \rto \left(\tilde{\GG}^{kl}-e^{A}e^{B}/n^2\right)/\sqrt{\left(v^{A}-(e^{A}/n)^2 \right) \left(v^{B}-(e^{B}/n)^2\right)}$ \Comment{center and normalize}

\EndFunction
\end{algorithmic}
\end{algorithm}

\clearpage

\begin{algorithm}
\caption{Compute the multiscale correlation map (i.e., all local generalized correlations) in $\mc{O}(n^2 \log n / T)$. Once the distances are sorted, the remaining algorithm runs in $\mc{O}(n^2)$. An important observation is that each product $a_{ij}b_{ij}$ is included in $\GG^{kl}$ if and only if $(k,l)$ satisfies $k\leq R(A_{\cdot j},i)$ and $l\leq R(B_{\cdot j},i)$, so it suffices to iterate through $a_{ij}b_{ij}$ for $i,j:=1,\ldots,n$, and add the product simultaneously to all $\GG^{kl}$ whose scales are no more than $(R(A_{\cdot j},i),R(B_{\cdot j},i))$. To achieve the above, we iterate through each product, add it to $\GG^{kl}$ at $(kl)=(R(A_{\cdot j},i),R(B_{\cdot j},i))$ only (so only one local scale is accessed for each operation); then add up adjacent $\GG^{kl}$ for $k,l=1,\ldots,n$. The same applies to all local covariances, variances, and expectations.
}
\label{alg:all_scales}
\begin{algorithmic}[1]
\Require A pair of distance matrices $(A, B) \in \Real^{n \times n} \times \Real^{n \times n}$.
\Ensure The multiscale correlation map $\mathcal{C} \in [-1,1]^{n \times n}$ for $k,l=1,\ldots,n$.
\Function{MGCAllLocal}{$A$, $B$}
\Linefor{$Z:=A,B$}{$\mathcal{E}^{Z}=\textsc{Sort}(Z)$}
\Linefor{$Z:=A,B$}{$Z=\textsc{Center}(Z)$}

\For{$i,j:=1,\ldots,n$} \Comment{iterate through all local scales to calculate each term}
\State $k \rto \mathcal{E}^{Z}_{ij}$
\State $l \rto \mathcal{E}^{Z}_{ij}$
\State $\tilde{\GG}^{kl} \rto \tilde{\GG}^{kl}+a_{ij}b_{ij}$
\State $v^{A}_{k} \rto v^{A}_{k}+a_{ij}^2$
\State $v^{B}_{l} \rto v^{B}_{l}+b_{ij}^2$
\State $e^{A}_{k} \rto e^{A}_{k}+a_{ij}$
\State $e^{B}_{l} \rto e^{B}_{l}+b_{ij}$
\EndFor

\For{$k:=1,\ldots,n-1$} \Comment{iterate through each scale again and add up adjacent terms}
\State $\tilde{\GG}^{1, k+1} \rto \tilde{\GG}^{1, k}+\tilde{\GG}^{1, k+1}$
\State $\tilde{\GG}^{k+1,1} \rto \tilde{\GG}^{k+1,1}+\tilde{\GG}^{k+1,1}$
\Linefor{$Z:=A,B$}{$v^{Z}_{k+1} \rto v^{Z}_{k}+v^{Z}_{k+1}$}
\Linefor{$Z:=A,B$}{$e^{Z}_{k+1} \rto e^{Z}_{k}+e^{Z}_{k+1}$}
\EndFor

\For{$k,l:=1,\ldots,n-1$}
\State $\tilde{\GG}^{k+1,l+1} \rto \tilde{\GG}^{k+1,l}+\tilde{\GG}^{k,l+1}+\tilde{\GG}^{k+1,l+1}-\tilde{\GG}^{k,l}$
\EndFor

\For{$k,l:=1,\ldots,n$}
\State $\GG^{kl} \rto \left(\tilde{\GG}^{kl}-e^{A}_{k}e^{B}_{l}/n^2\right)/\sqrt{\left(v^{A}_{k}-{e^{A}_{k}}^2/n^2\right) \left(v^{B}_{l}-{e^{B}_{l}}^2/n^2\right)}$
\EndFor
\EndFunction
\end{algorithmic}
\end{algorithm}

\clearpage

\section{Simulation Dependence Functions}
\label{appen:function}

This section provides the $20$ different dependency functions used in the simulations.  We used essentially the exact same relationships as previous publications to ensure a fair comparison \cite{SzekelyRizzoBakirov2007, SimonTibshirani2012, GorfineHellerHeller2012}. We only made changes to add white noise and a weight vector for higher dimensions, thereby making them more difficult, to better compare all methods throughout different dimensions and sample sizes. A few additional relationships are also included.

For each sample $\mb{x} \in \Real^{p}$, we denote $\mb{x}_{[d]}, d=1,\ldots,p$ as the $d^{th}$ dimension of the vector $\mb{x}$. For the purpose of high-dimensional simulations, $w \in \Real^{p}$ is a decaying vector with $w_{[d]}=1/d$ for each $d$, such that $w\TT \mb{x}$ is a weighted summation of all dimensions of $\mb{x}$.
Furthermore, $\mc{U}(a,b)$ denotes the uniform distribution on the interval $(a,b)$, $\mc{B}(p)$ denotes the Bernoulli distribution with probability $p$, $\mc{N}(\mu,{\Sigma})$ denotes the normal distribution with mean ${\mu}$ and covariance ${\Sigma}$,
$U$ and $V$ represent some auxiliary random variables, $\kappa$ is a scalar constant to control the noise level (which equals $1$ for one-dimensional simulations and $0$ otherwise), and $\epsilon$ is a white noise from independent standard normal distribution unless mentioned otherwise.

For all of the below equations, $(\mbx,\mby) \overset{iid}{\sim} F_{\mbx \mby} = F_{\mby|\mbx} F_\mbx$. For each relationship, we provide the space of $(\mbx,\mby)$, and define $F_{\mby|\mbX}$ and $F_\mbx$, as well as any additional auxiliary distributions.

\setcounter{equation}{0}
\begin{compactenum}
\item Linear $(\mbx,\mby) \in \Real^{p} \times \Real$,
\begin{align*}
\mbx &\sim \mc{U}(-1,1)^{p},\\
\mby &=w\TT \mbx+\kappa\epsilon.
\end{align*}
\item Exponential $(\mbx,\mby) \in \Real^{p} \times \Real$:
\begin{align*}
\mbx &\sim \mc{U}(0,3)^{p}, \\
\mby &=exp(w\TT \mbx)+10\kappa\epsilon.
\end{align*}
\item Cubic $(\mbx,\mby) \in \Real^{p} \times \Real$:
\begin{align*}
\mbx &\sim \mc{U}(-1,1)^{p}, \\
\mby &=128(w\TT \mbx-\tfrac{1}{3})^3+48(w\TT \mbx-\tfrac{1}{3})^2-12(w\TT \mbx-\tfrac{1}{3})+80\kappa\epsilon.
\end{align*}
\item Joint normal $(\mbx,\mby) \in \Real^{p} \times \Real^{p}$: Let $\rho=1/2p$, $I_{p}$ be the identity matrix of size $p \times p$, $J_{p}$ be the matrix of ones of size $p \times p$, and $\Sigma = \begin{bmatrix} I_{p}&\rho J_{p}\\ \rho J_{p}& (1+0.5\kappa) I_{p} \end{bmatrix}$. Then
\begin{align*}
(\mbx,\mby) &\sim \mc{N}(0, \Sigma).
\end{align*}
\item Step Function $(\mbx,\mby) \in \Real^{p} \times \Real$:
\begin{align*}
\mbx &\sim \mc{U}(-1,1)^{p},\\
\mby &=\mb{I}(w\TT \mbx>0)+\epsilon,
\end{align*}
where $\mb{I}$ is the indicator function, that is $\mb{I}(z)$ is unity whenever $z$ true, and zero otherwise.
\item Quadratic $(\mbx,\mby) \in \Real^{p} \times \Real$:
\begin{align*}
\mbx &\sim \mc{U}(-1,1)^{p},\\
\mby &=(w\TT \mbx)^2+0.5\kappa\epsilon.
\end{align*}
\item W Shape $(\mbx,\mby) \in \Real^{p} \times \Real$:  $U \sim \mc{U}(-1,1)^{p}$,
\begin{align*}
\mbx &\sim \mc{U}(-1,1)^{p},\\
\mby &=4\left[ \left( (w\TT \mbx)^2 - \tfrac{1}{2} \right)^2 + w\TT U/500 \right]+0.5\kappa\epsilon.
\end{align*}
\item Spiral $(\mbx,\mby) \in \Real^{p} \times \Real$: $U \sim \mc{U}(0,5)$, $\epsilon \sim \mc{N}(0, 1)$,
\begin{align*}
\mbx_{[d]}&=U \sin(\pi U)  \cos^{d}(\pi U) \mbox{ for $d=1,\ldots,p-1$},\\
\mbx_{[p]}&=U \cos^{p}(\pi U),\\
\mby&= U \sin(\pi U) +0.4 p\epsilon.
\end{align*}
\item Uncorrelated Bernoulli $(\mbx,\mby) \in \Real^{p} \times \Real$: $U \sim \mc{B}(0.5)$, $\epsilon_{1} \sim \mc{N}(0, I_{p})$, $\epsilon_{2} \sim \mc{N}(0, 1)$,
\begin{align*}
\mbx &\sim \mc{B}(0.5)^{p}+0.5\epsilon_{1},\\
\mby&=(2U-1)w\TT \mbx+0.5\epsilon_{2}.
\end{align*}
\item Logarithmic $(\mbx,\mby) \in \Real^{p} \times \Real^{p}$: $\epsilon \sim \mc{N}(0, I_{p})$
\begin{align*}
\mbx &\sim \mc{N}(0, I_{p}),\\
\mby_{[d]}&=2\log_{2}(|\mbx_{[d]}|)+3\kappa\epsilon_{[d]} \mbox{ for $d=1,\ldots,p$.}
\end{align*}
\item Fourth Root $(\mbx,\mby) \in \Real^{p} \times \Real$:
\begin{align*}
\mbx &\sim \mc{U}(-1,1)^{p},\\
\mby&=|w\TT \mbx|^\frac{1}{4}+\frac{\kappa}{4}\epsilon.
\end{align*}
\item Sine Period $4\pi$ $(\mbx,\mby) \in \Real^{p} \times \Real^{p}$: $U \sim \mc{U}(-1,1)$, $V \sim \mc{N}(0,1)^{p}$, $\theta=4\pi$,
\begin{align*}
\mbx_{[d]}&=U+0.02 p V_{[d]} \mbox{ for $d=1,\ldots,p$}, \\
\mby &=\sin ( \theta \mbx )+\kappa\epsilon.
\end{align*}
\item Sine Period $16\pi$ $(\mbx,\mby) \in \Real^{p} \times \Real^{p}$: Same as above except $\theta=16\pi$ and the noise on $\mby$ is changed to $0.5\kappa\epsilon$.
\item Square $(\mbx,\mby) \in \Real^{p} \times \Real^{p}$: Let $U \sim \mc{U}(-1,1)$, $V \sim \mc{U}(-1,1)$, $\epsilon \sim \mc{N}(0,1)^{p}$, $\theta=-\frac{\pi}{8}$. Then
\begin{align*}
\mbx_{[d]}&=U \cos\theta + V \sin\theta + 0.05 p\epsilon_{[d]},\\
\mby_{[d]}&=-U \sin\theta + V \cos\theta,
\end{align*}
for $d=1,\ldots,p$.
\item Two Parabolas $(\mbx,\mby) \in \Real^{p} \times \Real$: $\epsilon \sim \mc{U}(0,1)$, $U \sim \mc{B}(0.5)$,
\begin{align*}
\mbx &\sim \mc{U}(-1,1)^{p},\\
\mby &=\left( (w\TT \mbx)^2  + 2\kappa\epsilon\right) \cdot (U-\tfrac{1}{2}).
\end{align*}
\item Circle $(\mbx,\mby) \in \Real^{p} \times \Real$: $U \sim \mc{U}(-1,1)^{p}$, $\epsilon \sim \mc{N}(0, I_{p})$, $r=1$,
\begin{align*}
\mbx_{[d]}&=r \left(\sin(\pi U_{[d+1]})  \prod_{j=1}^{d} \cos(\pi U_{[j]})+0.4 \epsilon_{[d]}\right) \mbox{ for $d=1,\ldots,p-1$},\\
\mbx_{[p]}&=r \left(\prod_{j=1}^{p} \cos(\pi U_{[j]})+0.4 \epsilon_{[p]}\right),\\
\mby&= \sin(\pi U_{[1]}).
\end{align*}
\item Ellipse $(\mbx,\mby) \in \Real^{p} \times \Real$: Same as above except $r=5$.
\item Diamond $(\mbx,\mby) \in \Real^{p} \times \Real^{p}$: Same as  ``Square'' except $\theta=-\frac{\pi}{4}$.
\item Multiplicative Noise $(\mb{x},\mb{y}) \in \Real^{p} \times \Real^{p}$: $u \sim \mc{N}(0, I_{p})$,
\begin{align*}
\mb{x} &\sim \mc{N}(0, I_{p}),\\
\mb{y}_{[d]}&=u_{[d]}\mb{x}_{[d]} \mbox{  for $d=1,\ldots,p$.}
\end{align*}
\item Multimodal Independence $(\mbx,\mby) \in \Real^{p} \times \Real^{p}$: Let $U \sim \mc{N}(0,I_{p})$, $V \sim \mc{N}(0,I_{p})$, $U' \sim \mc{B}(0.5)^{p}$, $V' \sim \mc{B}(0.5)^{p}$. Then
\begin{align*}
\mbx&=U/3+2U'-1,\\
\mby&=V/3+2V'-1.
\end{align*}
\end{compactenum}

\begin{figure}[htbp]
\includegraphics[trim={5cm 1.5cm 4cm 0.5cm},clip, width=1.0\textwidth]{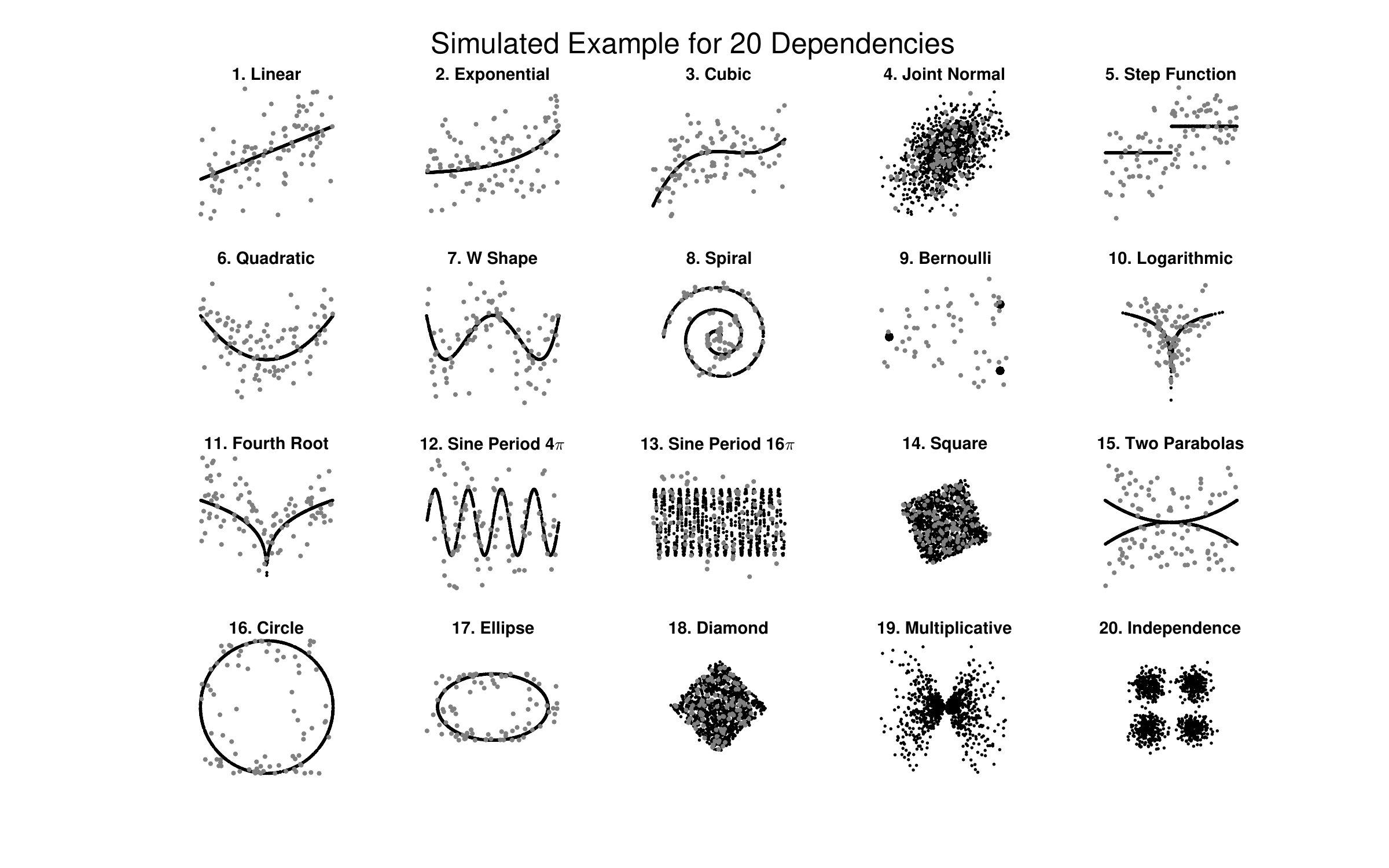}
\caption{Visualization of the $20$ dependencies at $p=q=1$. For each, $n=100$ points are sampled with noise ($\kappa=1$) to show the actual sample data used for one-dimensional relationships (gray dots). For comparison purposes, $n=1000$ points are sampled without noise ($\kappa=0$) to highlight each underlying dependency (black dots). Note that only black points are plotted for type 19 and 20, as they do not have the noise parameter $\kappa$.
}
\label{f:dependencies}
\end{figure}

For each distribution, $\mbx$ and $\mby$ are dependent except  (20); for some relationships (8,14,16-18) they are  independent upon conditioning on the respective auxiliary variables, while for others they are
 ``directly'' dependent.
A visualization of each dependency with $D=D_y=1$ is shown in Figure~\ref{f:dependencies}.

For the increasing dimension simulation in the main paper, we always set $\kappa=0$ and $n=100$, with $p$ increasing.  Note that $q=p$ for types  $4,10,12,13,14,18,19,20$, otherwise $q=1$.
The decaying vector $w$ is utilized for $p>1$ to make the high-dimensional relationships more difficult (otherwise, additional dimensions only add more signal).
For the one-dimensional simulations, we always set $p=q=1$, $\kappa=1$ and $n=100$.

\clearpage

\clearpage
\section{Supplementary Figures}
\label{appen:figs}

\begin{figure}[htbp]
\includegraphics[width=1.0\textwidth]{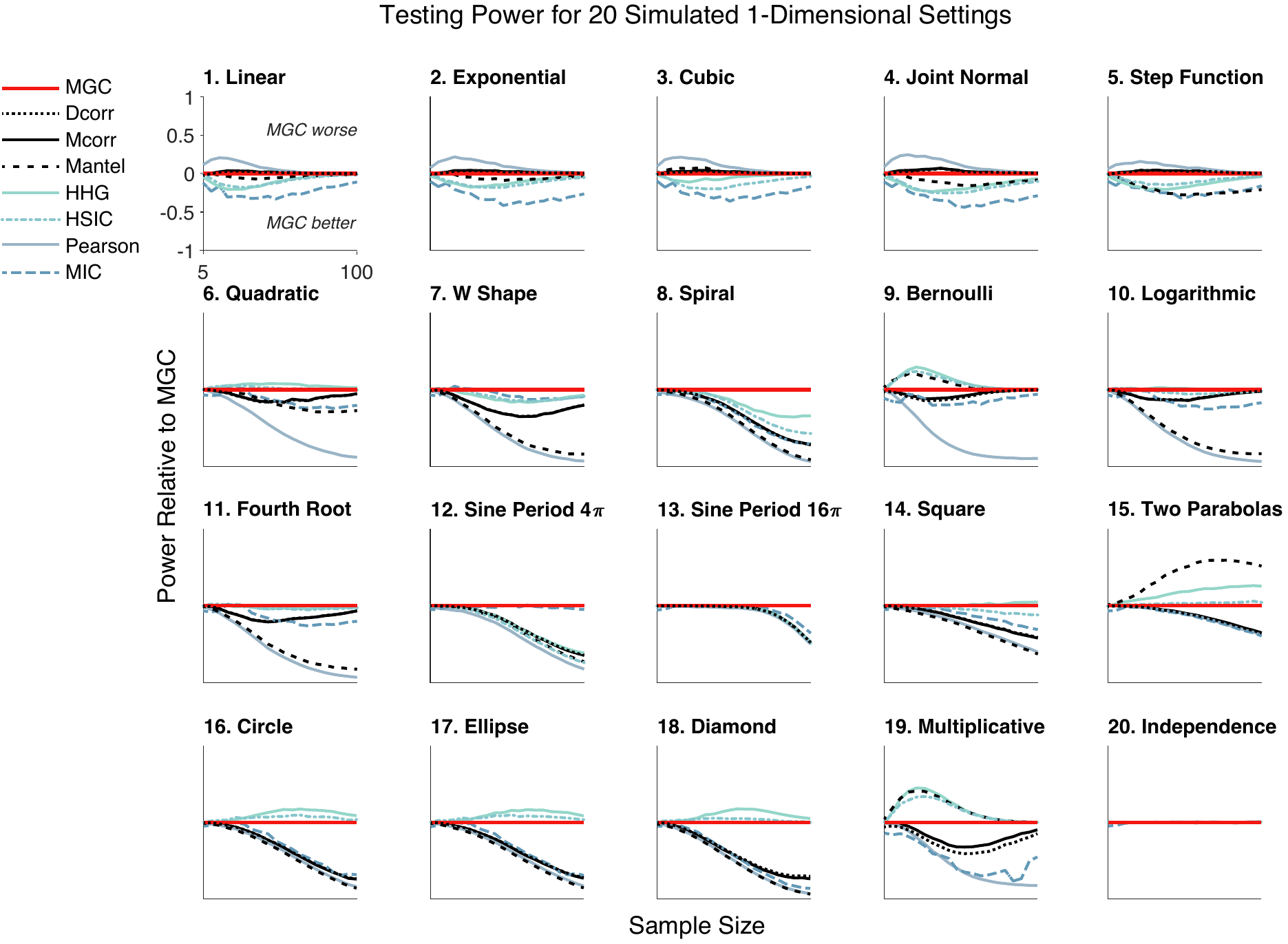}
\caption{
The same power plots as in Figure~\ref{f:nDAll}, except the $20$ dependencies are one-dimensional with noise, and the x-axis shows sample size increasing from $5$ to $100$.
Again, \Mgc~empirically achieves similar or better power than the previous state-of-the-art approaches on most problems. Note that \Mic~is included in 1D case; \RV~and \CCA~both equal \Pearson~in 1D; \Kendall~and \Spearman~are too similar to \Pearson~in power and thus omitted in plotting.}
\label{f:1DAll}
\end{figure}

\begin{figure}[htbp]
\includegraphics[width=1.0\textwidth,trim={0cm 0.5cm 3.5cm 0.5cm},clip]{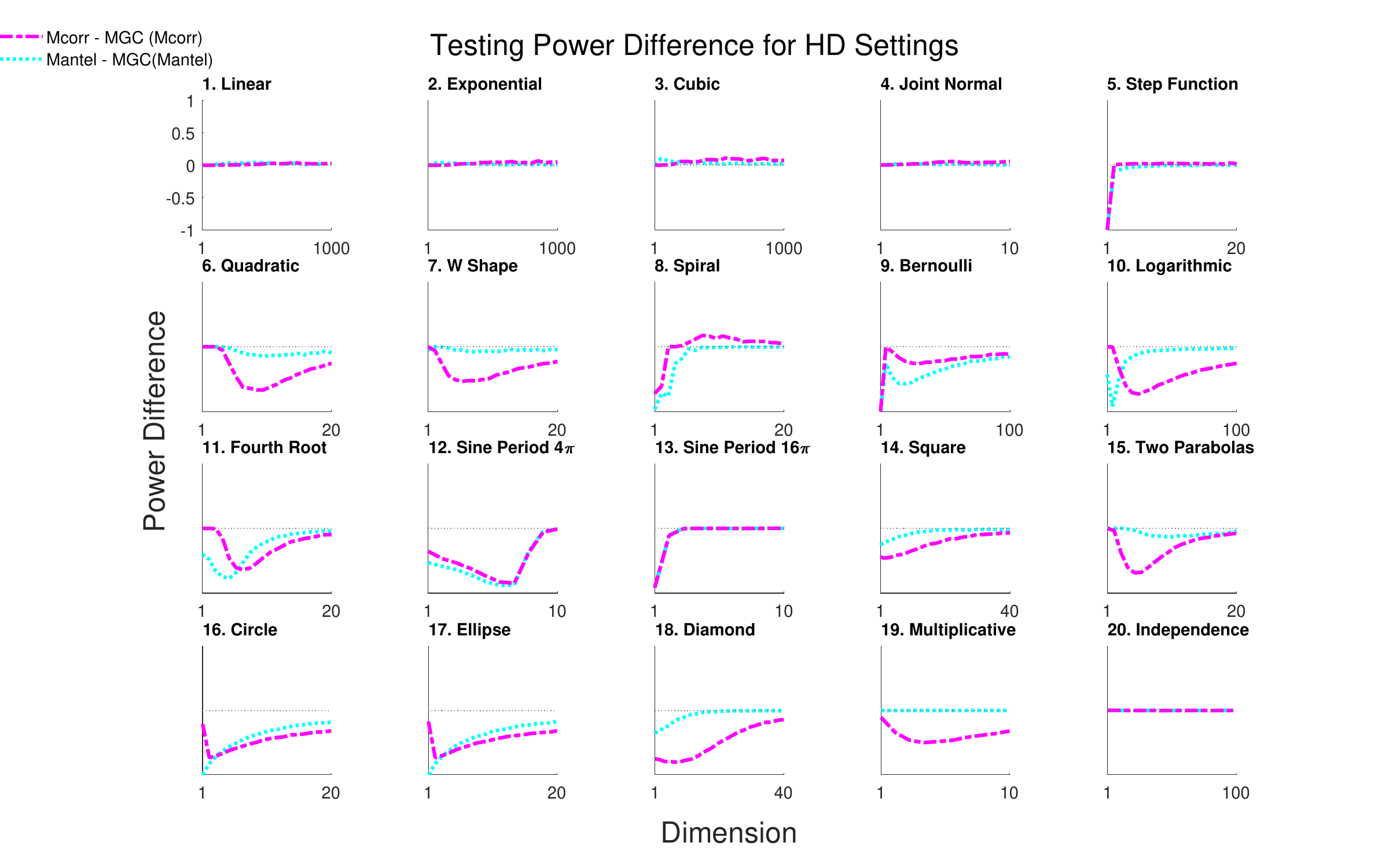}
\caption{The same set-ups as in Figure~\ref{f:nDAll}, comparing different \Mgc~implementations versus its global counterparts. The default \Mgc~builds upon \Mcorr~throughout the paper, and we further consider \Mgc~on \Mantel~to illustrate the generalization. The magenta line shows the power difference between \Mcorr~and \Mgc~, and the cyan line shows the power difference between \Mantel~and the  \Mgc~version of \Mantel. Indeed, \Mgc~is able to improve the global counterpart in testing power under nonlinear dependencies, and maintains similar power under linear and independent dependencies.}
\label{f:nDMantel}
\end{figure}

\begin{figure}[htbp]
\includegraphics[width=1.0\textwidth,trim={0cm 0.5cm 3.5cm 0.5cm},clip]{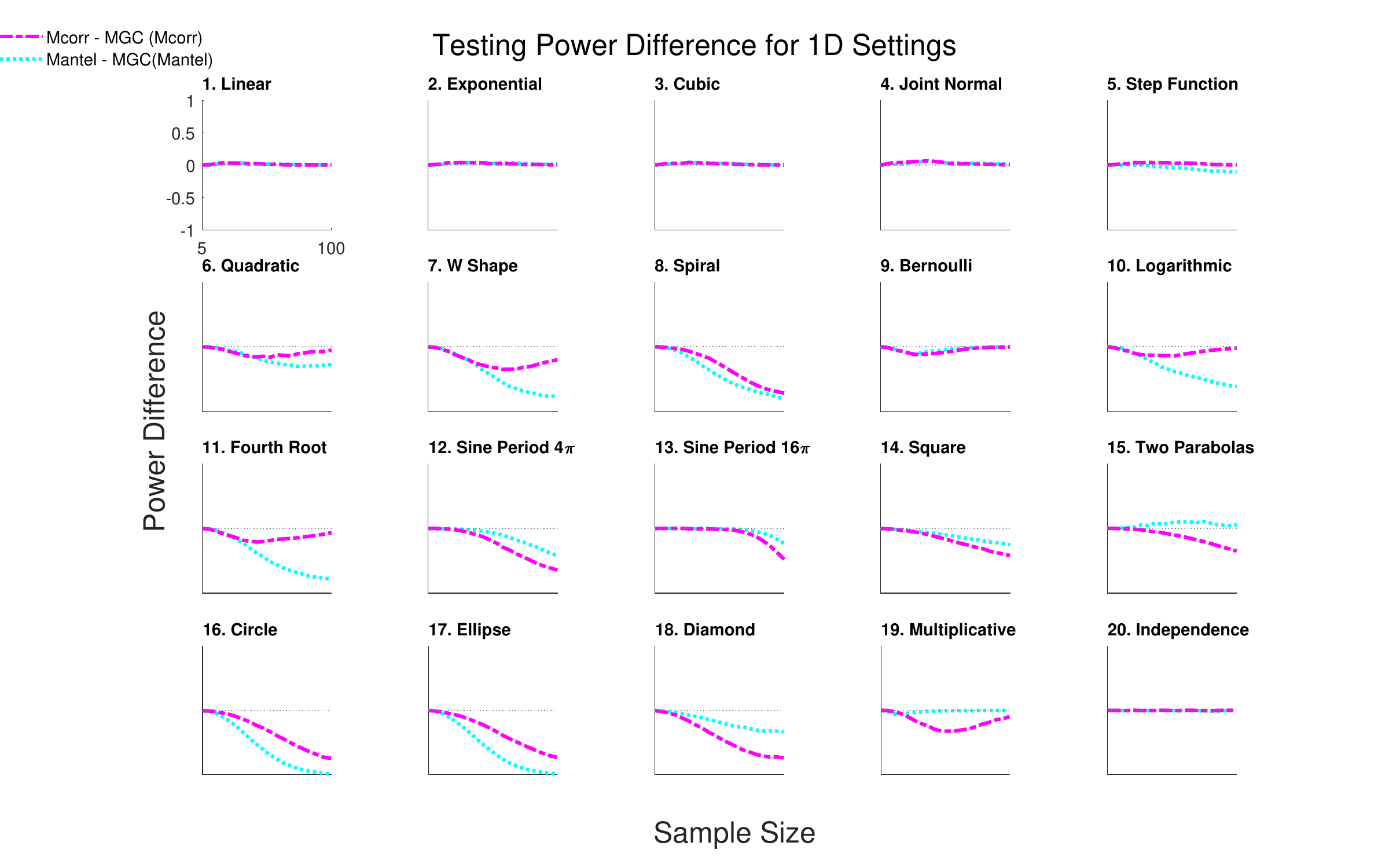}
\caption{The same power plots as in Figure~\ref{f:nDMantel}, except the $20$ dependencies are one-dimensional with noise, and the x-axis shows sample size increasing from $5$ to $100$.}
\label{f:1DMantel}
\end{figure}

\begin{figure}[htbp]
\includegraphics[width=1.0\textwidth,trim={5cm 1cm 4cm 0.5cm},clip]{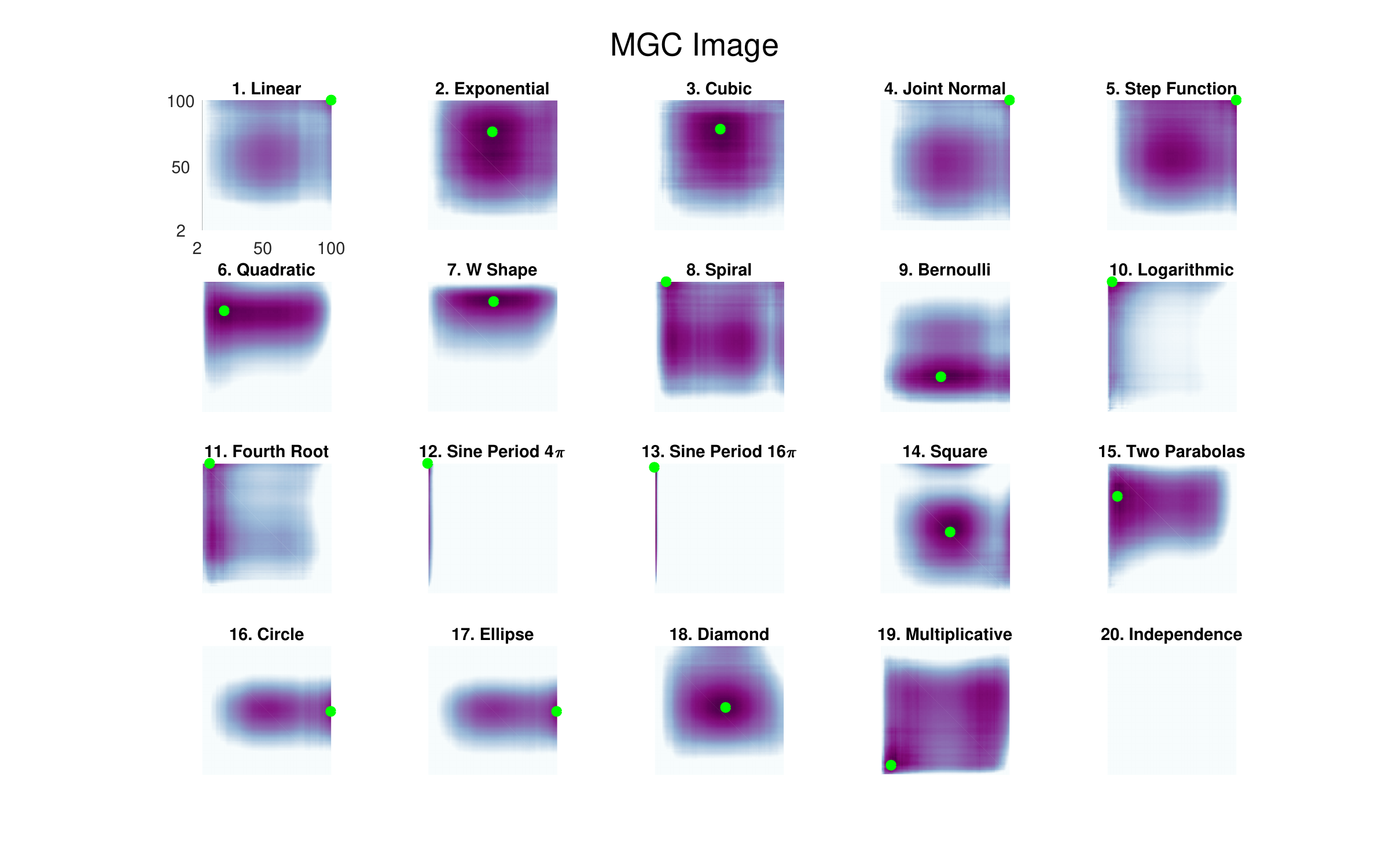}
\caption{The \Mgc-Map for the 20 panels for high-dimensional dependencies. For each simulation, the sample size is $100$, and the dimension is selected as the dimension such that \Mgc~has a testing power above $0.5$. It has similar behavior and interpretation as the 1-dimensional power maps in Figure~\ref{f:powermaps}, i.e., the linear relationships optimal scales are global, and similar dependencies share similar \Mgc-Maps.}
\label{f:powermaps1}
\end{figure}

\clearpage
\section{Real Data Processing}
\label{appen:real}

\subsection{Brain Activity vs Personality}
\label{app:personality}

This experiment investigates whether there is any dependency between resting brain activity and personality. Human personality has been intensively studied for many decades; the most widely used and studied approach is the NEO Personality Inventory-Revised the characterized personality along five dimensions \cite{Costa1992}.
This dataset consists of $42$ subjects, each with  $197$ time-steps of resting-state functional magnetic resonance activity (rs-fMRI) activity, as well as the subject's five-dimensional ``personality''. Adelstein et al. \cite{AdelsteinEtAl2011} were able to detect dependence between the activity of certain brain regions and dimensions of personality, but lacked the tools to test for dependence of whole brain activity against all five dimensions of personality.
For the five-factor personality modality, we  used the Euclidean distance. For the brain activity modality,
we derived the following comparison function. For each scan, (i) run Configurable Pipeline for the
 Analysis of Connectomes pipeline \cite{CPAC2015} to process the raw brain images yielding a parcellation into
197 regions of interest,
(ii) run a spectral analysis on each region and keep the power of band,
(iii) bandpass and normalize it to sum to one,
(iv) calculate the Kullback-Leibler divergence across regions to obtain a similarity matrix across comparing all regions.
Then, use the normalized Hellinger distance to compute distances between each subject.

\subsection{Brain Connectivity vs Creativity}
\label{app:creativity}

This experiment investigates whether there is any dependency between brain structural networks and creativity. Creativity has  been extensively studied in  psychology; the ``creativity composite index'' (CCI) is an index similar to an ``intelligence quotient'' but for creativity rather than intelligence \cite{Jung2009}.
This dataset consists of $109$ subjects, each with diffusion weighted MRI data as well as the subject's CCI.
Neural correlates of CCI have previously been investigated, though largely using structural MRI and cortical thickness \cite{Jung2009}.  Previously published results explored the relationship between graphs and  CCI \cite{Koutra15a}, but did not provide a valid test.
We used Euclidean distance to compare CCI values.
For the raw brain imaging data, we derived the following comparison function.  For each scan we estimated brain networks from diffusion and structural MRI data via  \Migraine, a pipeline for estimating brain networks from diffusion data \cite{GrayRoncal2013}.
We compute the distance between the graphs using the semi-parametric graph test statistic \cite{Sussman2013,ShenVogelsteinPriebe2016,Tang2016}, embedding each graph into two dimensions and aligning the embeddings via a Procrustes analysis.
\iffalse
\subsection{Brain Shape vs Depression}
\label{app:depression}

This experiment investigates whether there is any dependency between brain shape and depression.
This  dataset consists of $114$ subjects. Each subject has a structural MRI scan as well as a discrete variable indicating whether the subject is non-affected, high-risk, or clinically depressed.
Previous investigations have linked major depressive disorder to hippocampus shape \cite{ParkEtAl2008,PosenerEtAl2003}, though global tests were unable to detect a statistically significant dependence structure at the $\alpha=0.05$ level.
% From the MRI data, previous work extracted both the left and right hippocampi.
For the brain shape modality, we computed the distance utilizing  nonlinear landmark matching approach called Large Deformation Diffeomorphic Metric Mapping
 \cite{ParkEtAl2008,BegEtAl2005}.
For the depression variable, we used the Euclidean distance.
\fi

% \don{didn't understand that independence = two-sample}
\subsection{Proteins vs Cancer}
\label{app:cancer}
%  method establishment and data analysis

This experiment investigated whether there is any dependency between abundance levels of peptides in human plasma and the presence of cancers.  Selected Reaction Monitoring (SRM) is a targeted quantitative proteomics technique for measuring protein and peptide abundance in complicated biological samples \cite{PMID21248225}. In a previous study, we used SRM to identify $318$ peptides from
$33$ normal, $10$ pancreatic cancer, $28$ colorectal cancer, and $24$ ovarian cancer samples \cite{Wang2017}. Then, using other methods, we identifed three peptides that were implicated in ovarian cancer, and validated them as legitimate biomarkers with a follow-up experiment.
%
% SRM transition parameters (precursor ion M/z, fragmented ion M/z, collision energy, and dwell time) were optimized through using synthetic peptides. 200 $\mu$l human plasma from each individual (healthy or cancer patient) were depleted to remove top 14 high abundance proteins using a Seppro IgY14 LC10 column system purchased from Sigma-Aldrich (St. Louis, Missouri). Depleted human plasma samples were denatured, reduced, alkylated, trypsin-digested, and cleaned through hydrophobic solid phase extraction methods as previously described \cite{PMID21248225, Wang2017}. Target peptide abundances in each sample were determined using a set of SRM methods established using synthetic target peptides. The abundance of a target peptide was represented by the total area under the curve (AUC) of all its SRM transitions normalized to the total AUC of all SRM transitions from a heavy isotope labeled internal control peptide spiked with the same amount (3 femtomole) before the assay.
% Normalized AUC values for all the peptides in each sample were calculated to represent the relative abundances of the target peptides.
% %
% $33$ normal, $12$ pancreatic cancer, $29$ colorectal cancer, and $24$ ovarian cancer plasma samples were analyzed for the abundances of a panel of $318$ peptides. %, so the protein data is of size $318 \times 98$.

In this study, we performed the following five sets of tests on those data:
\begin{compactenum}
\item  ovarian vs.~normal for all proteins,
\item ovarian vs.~normal for each individual protein,
\item  pancreas vs.~normal for all proteins,
\item  pancreas vs.~all others for each individual protein,
\item pancreas vs.~normal for each individual protein.
\end{compactenum}
These tests are designed to first validate the \Mgc~method from ovarian cancer, then identify biomarkers unique to pancreatic cancer, that is, find a protein that is able to tell the difference between pancreas and normals, as well as pancreas vs all other cancers.
For each of the five tests, we create a binary label vector, with $1$ indicating the cancer type of interest for the corresponding subject, and $0$ otherwise. Then each algorithm is applied to each task.
For all tests we used Euclidean distances and the type $1$ error level is set to $\alpha=0.05$
The three test sets assessing individual proteins provide $318$ p-values; we used
% For the first two tests, the Euclidean distance of the peptides data is based on all proteins; for the last three screening tasks, the Euclidean distance matrix is based on each peptide, such that $318$ tests and p-values are reported. The type $1$ error level is set to $\alpha=0.05$, with
the Benjamini-Hochberg procedure \cite{Benjamini1995}  to control the false discovery rate. A summary of the results are reported in Table~\ref{t:real2}.

\begin{table*}[!ht]
\centering
\caption{Results for cancer peptide screening. The first two rows report the p-values for the tests of interest based on all peptides. The next four rows report the number of significant proteins from individual peptide tests; the Benjamini-Hochberg procedure is used to locate the significant peptides by controlling the false discovery rate at $0.05$.}
\label{t:real2}%
\begin{tabular}{|l|l||c|c|c|c|c|}
\hline
& Testing Pairs / Methods & Sample \Mgc & \Mantel & \Dcorr & \Mcorr & \Hhg \\
\hline
1 & Ovar vs. Norm: p-value & $\textbf{0.0001}$  & $\textbf{0.0001}$ & $\textbf{0.0001}$ & $\textbf{0.0001}$ & $\textbf{0.0001}$ \\
\hline
2 & Ovar vs. Norm: \# peptides  & $218$  & $190$ & $186$ & $178$ & $225$ \\
\hline \hline
3 & Pancr vs. Norm: p-value & $\mathbf{0.0082}$  & ${0.0685}$ & $0.0669$ & $0.0192$ & ${0.0328}$ \\
\hline
4 & Panc vs. Norm: \# peptides & $9$  & $7$ & $6$ & $7$ & $11$ \\
\hline
5 & Panc vs. All: \# peptides & $1$  & $0$ & $0$ & $0$ & $3$ \\
\hline
6 & \# peptides unique to Panc & $1$  & $0$ & $0$ & $0$ & $2$ \\
\hline
7 & \# false positives for Panc & $\mathbf{0}$  & n/a & n/a & n/a & $1$ \\
\hline
\end{tabular}
\end{table*}

All methods are able to successfully detect a dependence between peptide abundances in ovarian cancer samples versus normal samples (Table \ref{t:real2}, line 1). This is likely because there are so many individual peptides that have different abundance distributions between ovarian and normal samples (Table \ref{t:real2}, line 2).  Nonetheless, \Mgc~identified more putative biomarkers than any of the other methods.  While we have not checked all of them with subsequent experiments to identify potential false positives, we do know from previous experiments that three peptides in particular are effective biomarkers.
%
% For the first two tests, \Mgc~reports as low or lower p-value than others.
%
% For the, and is the only method other than \Hhg~that can reveal useful proteins for pancreas versus normal persons and also pancreas versus all other cancer types.
%
% For the third test, i.e., feature screening for ovarian vs. normal, previously we validated three peptides by other techniques: peptidyl-prolyl cis-trans isomerase A, dynamin-3 isoform b, peptidyl-prolyl cis-trans isomerase A.
% It turns out that
All three peptides have p-value $\approx 0$ for all methods including \Mgc, that is, they are all correctly identified as significant.
% However, there are a large number of significant peptides identified for each method as shown in Table~\ref{t:real2} third column.
However, by ranking the peptides based on the actual test statistic of each peptide,
% the three peptides have rank $101,69,15$ by \Mgc, $143,92,33$ by \Mantel, $178,95,36$ by \Dcorr~and \Mcorr, $88,111,12$ by \Hhg.
\Mgc~is the method that ranks the three known biomarkers the lowest, suggesting that it is the least likely to falsely identify peptides.

We then investigated the pancreatic samples in an effort to identify biomarkers that are unique to pancreas. We first checked whether the methods could identify a difference using all the peptides.  Indeed, three of the five methods found a dependence at the $0.05$ level, with Sample \Mgc~obtaining the lowest p-value (Table \ref{t:real2}, line 3).  We then investigated how many individual peptides the methods identified; all of them found 6 to 11 peptides with a significant difference between pancreatic and normal samples (Table \ref{t:real2}, line 4).  Because we were interested in identifying peptides that were uniquely useful for pancreatic cancer, we then compared pancreatic samples to all others.  Only \Mgc, \Hsic, and \Hhg~identified peptides that expressed different abundances in this more challenging case (Table \ref{t:real2}, line 5).  To identify peptides that are unique to pancreatic cancer, we looked at the set of peptides that were both different from normals and different from all non-pancreatic cancer samples (Table \ref{t:real2}, line 6).
%\Mgc~and \Hhg~found $1$ and $2$ peptides satisfying those conditions, respectively (Table \ref{t:real2}, line 6).
%For the last two tests exploring pancreatic, b
All three method reveal the same unique protein for pancreas: neurogranin. \Hsic~identifies another peptide (tropomyosin alpha-3 chain isoform 4), and \Hhg~identifies a third peptide (fibrinogen-like protein 1 precursor). However, fibrinogen-like protein 1 precursor is not significant for p-value testing between pancreatic and normal subjects. On the other hand, tropomyosin is a ubiquitously expressed protein, since normal tissues and other cancers will also express tropomyosin and leak it into blood, whereas neurogranin is exclusively expressed only in brain tissues. Moreover, there exists strong evidence of tropomyosin 3 upregulated in other cancers \cite{TPM1,TPM2,TPM3,TPM4}. Therefore, initial literature search suggests that tropomyosin is likely falsely identified by \Hhg~and less useful as a  pancreatic cancer marker, meaning that only \Mgc~identified putative pancreatic cancer biomarkers without also identifying likely false positives.

Furthermore, although neurogranin is not identified by other methods, it is always the most dependent peptide in all methods except \Mic. Namely, all of \Pearson, \Dcorr, \Mcorr, \Mantel, \Hhg, \Hsic, and \Mgc~rank neurogranin as the most significant protein by p-value; the only difference is that the p-values are not significant enough for other methods after multiple testing adjustment. Also, the three peptides identified by \Hhg~are also the top three in \Mgc; and if we further investigate the top three peptides in all methods, they always come from these three peptides, and another peptide (mitogen-activated protein kinase); the only exception is \Mic, whose top three peptides do not coincide with all other correlation measures, which suggests it may detect too many false positives. Along with the classification result showing that neurogranin along has the best classification error, this experiment strongly indicate that \Mgc, \Hsic, \Hhg~are the top methods in dependency testing, able to amplify the signal, and do not detect false signals.

\subsection{\Mgc~Does Not Inflate False Positive Rates in Screening}

\begin{figure}[htbp]
\centering
\includegraphics[width=0.5\textwidth,trim={0cm 0cm 0cm 0cm},clip]{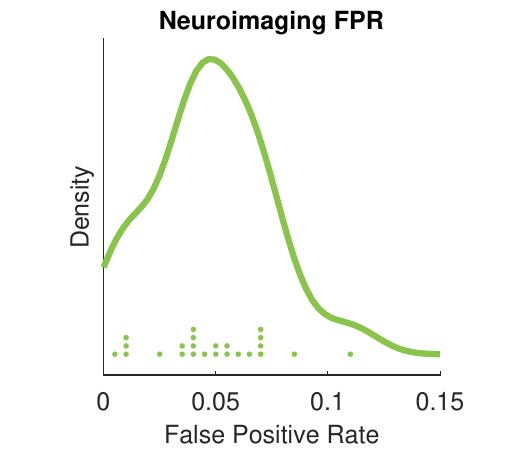}
\caption{We demonstrate that \Mgc~is a valid test that does not inflate the false positives in screening and variable selection. This figure shows the density estimate for the false positive rates of applying \Mgc~to  select the ``falsely significant" brain regions versus independent noise experiments; dots indicate the false positive rate of each experiment. The mean $\pm$ standard deviation is $0.0538 \pm 0.0394$.}
\label{f:fpr}
\end{figure}

%In the previous  experiment, \Mgc~effectively selects the true positives in a cancer proteomics screening.
In this final experiment, we empirically determine that \Mgc~does not inflate false positive rates via a neuroimaging screening.
% demonstrate that \Mgc~does not have high false negative rates, in the final experiment we empirically test whether \Mgc~inflates false positive rates.
To do so, we extend the work of Eklund et al. \cite{EklundKnutsson2012,Eklund2015},
where a number of parametric methods are shown to largely inflate the false positives. Specifically, we applied \Mgc~to test whether there is any dependency between brain voxel activities and random numbers.
For each brain region, \Mgc~attempts to test the following hypothesis: Is activity of a  brain region independent of the time-varying stimuli?
Any region that is selected as significant is a false positive by construction.  By testing each brain region separately, \Mgc~provides a distribution of false positive rates.  If \Mgc~is valid, the resulting distribution should be centered around the significance level, which is set at $0.05$ for these experiments.
We considered $25$ resting state fMRI experiments from the $1$,$000$ Functional Connectomes Project  consisting of a total of $1$,$583$ subjects \cite{biswal2010toward}.
Figure~\ref{f:fpr} shows the false positive rates of  \Mgc~for each dataset, which are centered around the critical level $0.05$, as it should be.
% \don{is the above really true?}
In contrast, many standard parametric methods for fMRI analysis, such as generalized linear models, can significantly increase the false positive rates, depending on the data and pre-processing details \cite{EklundKnutsson2012,Eklund2015}. Moreover, even the proposed solutions to those issues make linearity assumptions, thereby limiting detection to only a small subset of possible dependence functions.

\subsection{Running Time Report}
\label{appen:time}
Here we list the actual running time of \Mgc~versus other methods for testing on the real data, based on a modern desktop with a six core I7-6850K CPU and 32GB memory on Matlab 2017a on Windows 10. The first two experiments are timed based on $1000$ permutations, while the screening experiment is timed without permutation, i.e., compute the test statistic only. \Pearson~runs the fastest, trailed by \Mic~and then \Dcorr. \Pearson~and \Mic~are only possible to run in the screening experiment, as the other two experiments are multivariate.  The running time of \Mgc~is a constant times (about $10$) higher than that of \Dcorr, and \Hhg~is implemented in a running time of $O(n^3)$ and thus significantly slower. %Note that a faster implementation is available in $O(n^2 \log n)$ for \Hhg, but here we use the slower version to better benchmark the time difference of different methods.

\begin{table*}[!ht]
%\centering
\caption{The Actual Testing Time (in seconds) on Real Data.}
\begin{tabular}{*4l}
\toprule
Data & Personality & Creativity & Screening   \\
\midrule
%  Oracle \Mgc  & \textbf{50}  & 60 & \textbf{70} & \textbf{135} \\
\Mgc  & 2.5  & 7.5 & 1.9  \\
  \Dcorr & 0.2  & 0.4 & 0.18 \\
  \Hsic & 0.5  & 1.7 & 0.23 \\
 \Hhg & 6.3  & 53.4 & 12.3   \\
 \Pearson & NA  & NA & 0.03  \\
 \Mic & NA  & NA & 0.1   \\
\bottomrule
\end{tabular}
\end{table*}

\end{document}